# Prediction, Generation of WWTPs microbiome community structures and Clustering of WWTPs various feature attributes using DE-BP neural network model, SiTime-GAN model and DPNG-EPMC ensemble clustering algorithm with modulation of microbial ecosystem health

*Mingzhi Dai[1,2,3], Weiwei Cai[4], Xiang Feng[3], Huiqun Yu[3], Weibin Guo[3] and Miao Guo[2]\**

[1] School of Aeronautics and Astronautics, Shanghai Jiao Tong University, Shanghai No.800 Dongchuan Road, Minhang District, Shanghai, 200240, China

[2] Department of Engineering, Faculty of Natural, Mathematical & Engineering Sciences, King's College London, Stand Building, Strand, WC2R 2LS, London, United Kingdom

[3] Department of Information Science and Engineering, East China University of Science and Technology, Shanghai Engineering Research Center of Smart Energy, Shanghai, Number 130 Meilong Road, Xuhui District, Shanghai, 200237, China

[4] School of Environment, Beijing Jiaotong University, No. 3 Shangyuan Village, Haidian District, Beijing, 100044, China

**Corresponding Author E-mail**: miao.guo@kcl.ac.uk

**Abstract**

Microbiomes not only underpin Earth's biogeochemical cycles but also play crucial roles in both engineered and natural ecosystems, such as the soil, wastewater treatment, and the human gut. However, microbiome engineering faces significant obstacles to surmount to deliver the desired improvements in microbiome control. Here, we use the backpropagation neural network (BPNN), optimized through differential evolution (DE-BP), to predict the microbial composition of activated sludge (AS) systems collected from wastewater treatment plants (WWTPs) located worldwide. Furthermore, we introduce a novel clustering algorithm termed Directional Position Nonlinear Emotional Preference Migration Behavior Clustering (DPNG-EPMC). This method is applied to conduct a clustering analysis of WWTPs across various feature attributes. Finally, we employ the Similar Time Generative Adversarial Networks (SiTime-GAN), to synthesize novel microbial compositions and feature attributes data. As a result, we demonstrate that the DE-BP model can provide superior predictions of the microbial composition. Additionally, we show that the DPNG-EPMC can be applied to the analysis of WWTPs under various feature attributes. Finally, we demonstrate that the SiTime-GAN model can generate valuable incremental synthetic data. Our results, obtained through predicting the microbial community and conducting analysis of WWTPs under various feature attributes, develop an understanding of the factors influencing AS communities.

# 1. Introduction

The activated sludge (AS) system is one of the commonly used wastewater treatment methods in wastewater treatment plants (WWTPs) [1-3]. The basic principle of this system is to mix wastewater with already activated microorganisms to break down and remove organic substances and pollutants from the wastewater. This method has been widely applied in urban and industrial wastewater treatment plants for over a century. It effectively removes organic matter, nitrogen, phosphorus, and other pollutants, ensuring the wastewater meets environmental discharge standards or water quality requirements for reuse [4-5]. Additionally, the AS system has advantages such as strong adaptability, high treatment efficiency, and relatively easy operation and management, making it a preferred choice among engineers and environmental agencies [6].

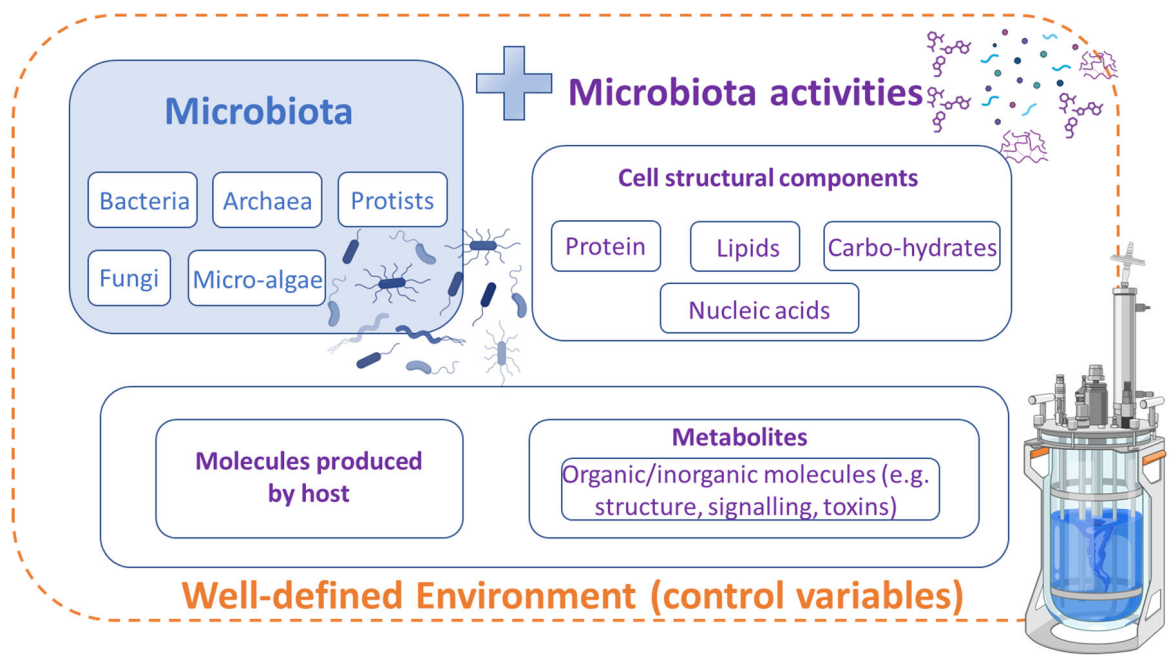

**Figure 1.** The collection of microorganisms and activities, and the molecules produced within a well-defined environment

The treatment standards and factors for wastewater treatment plants (WWTPs) can vary depending on the regulations, environmental conditions, and wastewater types in different countries and regions. However, in general, AS-core WWTPs are required to ensure that the discharged water quality complies with water quality standards set by national or local governments. These standards typically include limits on various pollutants such as suspended solids [7], chemical oxygen demand (COD) [8], biochemical oxygen demand (BOD) [9], nitrogen (N) [10], phosphorus (P) [11], heavy metals [12], and more. Clearly, the intrinsic core in AS was the microbiome, which underpins the biological treatment efficiency of WWTPs and determines the effluent quality. Wastewater treatment plants need to select appropriate treatment processes to ensure efficient removal of pollutants. Common treatment processes include activated sludge, biofilters, chemical precipitation, and membrane separation [13-15]. Additionally, in various scientific disciplines and microbiome systems, researchers have applied several ecological concepts to investigate questions related to the impact of microbiome modulation on both microbiome composition and ecosystem function. The microbiome comprises a collection of microorganisms and their activities [16-18], involving a spectrum of molecules produced within a well-defined environment (as shown in **Figure 1**). Understanding the macrobiotics data in AS would be the key to regulating the WWTPs treatment efficiency, thereby a burst of emerging omics tools ambitiously aims to open the black box. While metagenomic technologies have provided deeper insights into microbial biodiversity and distribution within diverse microbiome ecosystems (as shown in **Figure 2**), there remains a wealth of hidden information and patterns within high-dimensional data [19] that have yet to be uncovered. Notably, by exploring these underlying patterns, we can achieve the following objectives: 1) predict the interactions between the microbiome and the environment (or host) as well as the composition of the microbiome. 2) classify microbial features and identify the 'core' microbial community to assess the stability and health of ecosystems. 3) understand the spatial and temporal characteristics of the microbiome to further optimize biotechnological applications, such as wastewater treatment (WWT) [20-22].

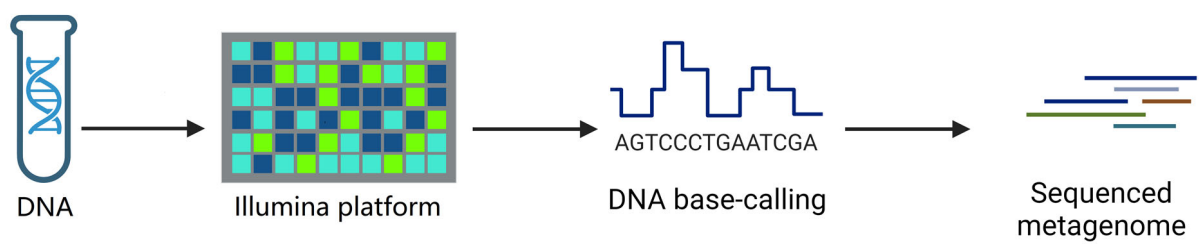

**Figure 2.** The Metagenomic techniques to determine the microbial community structure

Artificial intelligence (AI)-based modeling approaches hold the potential to tackle the aforementioned challenges. While there have been ongoing efforts to adapt and implement machine learning (ML) algorithms in microbiome studies, fully harnessing the power of AI for accelerating exciting discoveries necessitates domain knowledge for algorithm selection and hyperparameter tuning. A promising avenue to overcome this knowledge barrier is the development of ML tools [23-24] capable of autonomously managing data [25], designing ML models [26], and optimizing model architecture and parameters with minimal intervention from end-users. Kinetic models rooted in mechanisms, such as the Monod equation [27], Lotka - Volterra model [28-29], and individual-based dynamic model [30], can predict the makeup of microbial communities by accounting for precise growth and interaction mechanisms in predefined conditions. However, some dynamic models [31-32] can be highly complex, demanding sophisticated mathematical methods and computational techniques for their solutions. This complexity can make the construction and application of the models intricate, requiring specialized knowledge and tools. Multiple linear regression models can utilize various environmental factors to predict the structure of microbial communities. Previous studies have employed this method [33-35] to predict the composition of fungal communities in microbial communities based on factors such as C and N concentrations, pH, Mean Annual Temperature (MAT), Mean Annual Precipitation (MAP), and Net Primary Productivity (NPP) [22]. However, they ignored several key factors in predicting WWTP microbial communities, such as environmental and geographical features, influent features, effluent features, reactor operation features, as well as features like removal rate for TP (total phosphorus), TN (total nitrogen), COD (chemical oxygen demand) BOD (biochemical oxygen demand). Therefore, incorporating more predictable regularities of microbes is crucial for gaining a deeper understanding of the composition of microbial communities.

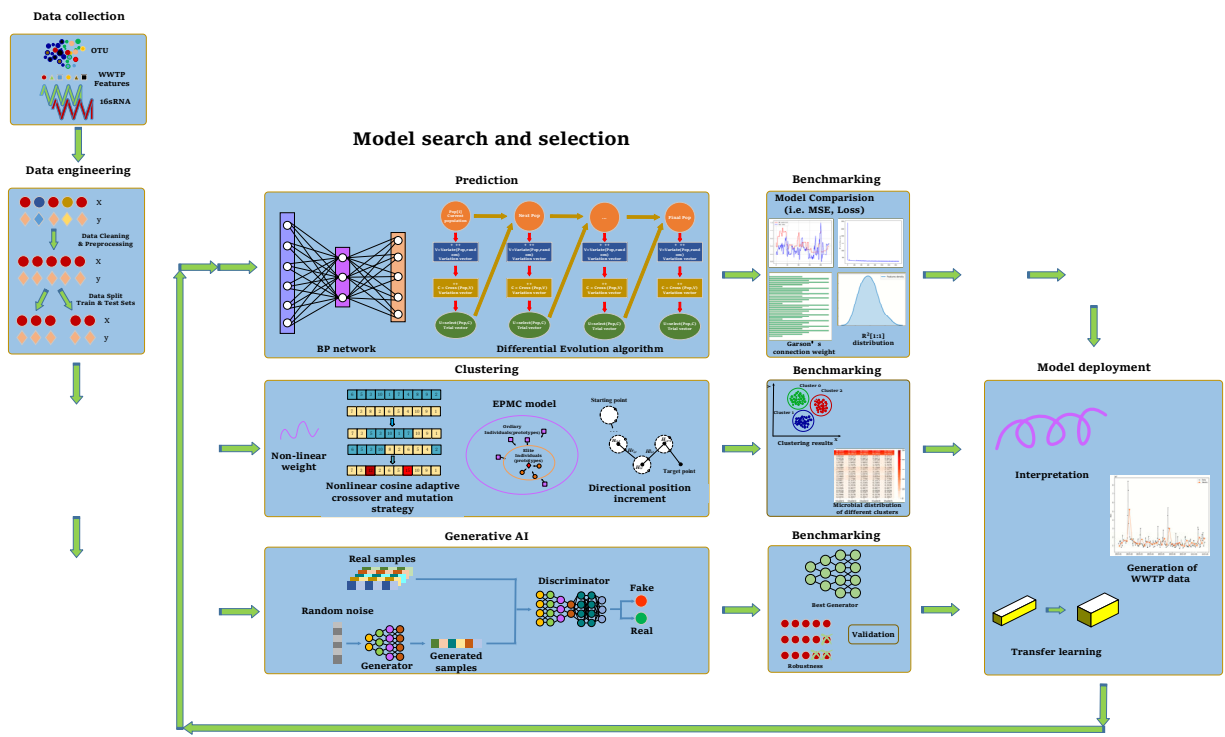

**Figure 3.** The whole framework of WWTPs microbial composition prediction and WWTPs clustering

In general, we employed the DE-BP model for the prediction of global wastewater treatment plant (WWTP) microbial community structures, utilized the DPNG-EPMC algorithm to cluster WWTPs according to various feature attributes, and introduced the GANs model to generate new data of various feature attributes and microbial community compositions of WWTPs for further prediction. The whole framework of WWTPs microbial composition prediction and WWTPs clustering is shown in **Figure 3**. The integrated pipeline includes data preprocessing, three different models for prediction, clustering, and generation, and related benchmarking of models. In prediction, the trained DE-BP model is automatically iterated for parameter optimization and prediction, then we used the Garson's connection weight to analyze the impact of various WWTPs characteristic factors and 1:1 observed-predicted distributions to measure the predictive accuracy; in clustering, the novel DPNG-EPMC algorithm is applied to the clustering analysis of WWTPs under different feature attributes, then we generate heatmaps to depict the proportion of different microbial community in various WWTPs feature attributes; in generation, the GANs model generates various feature attributes and microbial community composition data, and we put the new data back to train the DE-BP model in order to achieve better predictive results.

## 2. Background and method

In this section, we present the relevant models and theoretical foundations of the DE-BP neural network, the swarm intelligence ensemble DPNG-EPMC algorithm, and the generative SiTime-GAN model for predicting microbial community structure in WWTPs, clustering WWTPs under various feature attributes, and generating substantial data on different attributes of WWTPs to train the DE-BP model further, respectively.

### 2.1. The DE-BP model for WWTPs prediction [49-50]

The Differential Evolution algorithm is a stochastic heuristic search technique that can be viewed from an engineering perspective as an adaptive iterative optimization process, as shown in **Supplementary Figure S2 and Supplementary Note 2** [45-48]. The steps of the Differential Evolution algorithm are roughly as follows:

Randomly select two individuals $x_{r1}^t$ and $x_{r2}^t$ in the population, then perform the scaling operation with the parameter $F$ on the vector difference of them, and synthesize it with the individual $x_i^t$ to obtain the mutated individual $va_i^{t+1}$, as calculated by Equation 1:

$$va_i^{t+1} = x_i^t + F \times (x_{r1}^t - x_{r2}^t) \quad (1)$$

$$\varphi = e^{1-\frac{G_m}{G_m+1-G}} \quad (2)$$

$$F = F_0 \cdot 2^\varphi \quad (3)$$

where $x_{r1}^t$ and $x_{r2}^t$ are the randomly selected individuals in the t-th iteration, $x_i^t$ represents the individual to be mutated, and F is the adaptive scaling factor. $G_m$ and $G$ in Equation 2 represent the maximum number of evolutions and the current evolution number, respectively. Besides, $va_i^{t+1}$ represents the mutant individual $i$ of the $(t+1)$-th iteration.

Next, the model employs the binomial crossover to further enhance the diversity of the population. The crossover operations are performed on the t-th generation population $x_{i,j}^t$ and its variant intermediate $va_{i,j}^{t+1}$. The cross condition is shown in Equation 4:

$$u_{i,j}^{t+1} = \begin{cases} va_{i,j}^{t+1}, & \text{if } rand(0,1) \leq CR \text{ or } j = j_{rand} \\ x_{i,j}^t, & \text{if } rand(0,1) > CR \end{cases} \quad (4)$$

where $va_{i,j}^{t+1}$ represents the intermediate individual variated by the individual $x_{i,j}^t$, and $CR$ denotes the crossover probability. In order to ensure that each chromosome (individual) has at least one gene of the variant chromosome reserved for the next generation, we randomly select a gene of $va_{i,j}^{t+1}$ as the $j_{rand}$ allele of chromosome $u_{i,j}^{t+1}$, where $j_{rand}$ represents a random integer in the interval [1, D]. Then, the subsequent operation utilizes the crossover probability CR to determine each allele of individual $u_{i,j}^{t+1}$.

Through the mutation and binomial crossover operations, a temporary individual is generated. Finally, a greedy algorithm is utilized to select these two individuals to produce a new generation. The selection process is shown in Equation 5:

$$x_{i,j}^{t+1} = \begin{cases} u_{i,j}^{t+1}, & \text{if } fit(u_{i,j}^{t+1}) \leq fit(x_{i,j}^t) \\ x_{i,j}^t, & \text{otherwise} \end{cases} \quad (5)$$

where fit represents the fitness function, which can measure the quality of individuals in the population, then the individual continues to evolve until the termination condition is met.

As depicted in Supplementary **Figure S1 and Supplementary Note 1** [36-44], the input layer of BPNN is the various environmental factors of WWTPs, which is replaced by $M$ nodes; the output layer is different microbial community structures, which is replaced by $N$ nodes (varying at levels Phylum, Class, and Order); and the number of nodes in the hidden layer, $\sqrt{M+N}+\alpha$, is calculated by the empirical formula. Once initial parameters are chosen, the gradient descent algorithm treats these initial parameter values as a starting point for parameter optimization and subsequent updates. However, due to the potentially suboptimal nature of randomly initialized parameters, the accuracy of the trained model could be significantly influenced.

As a heuristic algorithm, Differential Evolution (DE) exhibits strong global search capabilities, making it a suitable candidate for addressing the optimization problem of neural network parameters. In the DE-BP neural network, the parameters serve as the decision variables of the problem, while the accuracy of the model functions as the objective function. In the context of microbiome community prediction tasks, the improved DE-BP algorithm can be effectively employed for optimizing parameters during neural network training. The flowchart of DE-BP neural network and the DE-BP neural network for WWTPs prediction are described in **Supplementary Figure S3 and Supplementary Note 3** [49-50], respectively.

**2.2. The DPNG-EPMC algorithm for WWTPs clustering**

2.2.1. Basic definition for clustering

**Definite1** (*Individual*) The individuals in the population can be represented by a matrix $H=l*k$, as shown in Equation 6, where $l$ represents the number of clusters and $k$ represents the dimension of the data set. Each row of the matrix represents a cluster center of the dataset, and the columns represent the same characteristics of the dataset.

$$H = \begin{bmatrix} h_{11} & h_{12} & ... & h_{1k} \\ h_{21} & h_{22} & ... & h_{2k} \\ ... & ... & ... & ... \\ h_{l1} & h_{l2} & ... & h_{lk} \end{bmatrix} \in R \qquad (6)$$

During the clustering, all objects are assigned to different clusters. The Euclidean distance is a commonly utilized distance metric to measure the effect of clustering.

**Definite 2** (*Fitness value*) To measure the quality of individuals in the population, we define a standard function for evaluation. The fitness of the individual is shown in Equation 7,

$$fit(l, H) = \frac{l \times \sum_{i=1}^{n} \|inst_i - inst_{label}\|}{\min\|h_m - h_n\|} \qquad (7)$$

where $H$ represents the matrix of individuals, and $l$ represents the number of clusters. In Equation 7, the numerator represents the instance cohesion of each group, where $inst_i$ donates an instance $i$ in the dataset, and $inst_{label}$ represents the label of the object $i$. The denominator represents the dispersion between different clusters, and $h_m$ and $h_n$ represent two cluster centers (rows) in the individual $H$. It can be concluded from Equation 7 that the larger the denominator or smaller the numerator, the smaller the fitness. That is, the better the individual $i$.

The clustering process can be expressed as: given a dataset $R = \{inst_1, inst_2, ..., inst_i\}$ containing i instances, the model divide the dataset $R$ into l clusters $R = \{r_1, r_2, ..., r_l\}$. At this point, the cluster centers $C = \{c_1, c_2, ..., c_l\}$ will be generated, which is a matrix composed of $l$ cluster centers. Therefore, Equation 7 can be used to describe the quality of individuals, and the goal of optimization is to find the best individual $H_{op}$. That is, Equation 7 reaches the minimum. The detailed description of the WWTPs clustering is in **Supplementary Figure S4 and Supplementary Note 4.**

### 2.1.2. The mathematical method in DPNG-EPMC algorithm

**Emotional preference and migratory behavior clustering (EPMC) model [51]**

First, the EPMC finds the optimal through iterative information learning and update between individuals. The model regards each individual as a habitat, and the migration rate of the habitat represents the probability of the individual being learned, which can provide more opportunities for high-quality individuals. Firstly, individuals are sorted according to fitness value $fit(H_1) < fit(H_2) < \cdots < fit(H_{NP})$. The weight of an individual $i$ can be represented by Equation 8 and the probability of individual $i$ being learned is defined as Equation 9.

$$\phi_i = \frac{NP-i+1}{NP} \tag{8}$$

$$P_i = \frac{\phi_i}{\sum_{i=1}^{NP} \phi_i} \tag{9}$$

Where $NP$ denotes the scale of the population.

If an individual in the population satisfies $P_1+P_2+\cdots+P_{m-1} < P_{rand} < P_1+P_2+\cdots+P_m$, then the individual $H_{m,t}$ is selected as the learning object, where $P_{rand}$ is a random number in the interval [0,1].

Second, the neighbor set of individual $i$ is defined as $N = \{H_i | i \in [i-n, i+n]\}$. The model utilizes the emotional preference matrix $feel_{i,j}$ to describe the credibility between individuals, as shown in Equation 10. After learning and update, the credibility of an individual increases or decreases. At the same time, the probability of individual $H_i$ being learned by neighbors is defined as Equation 11.

$$feel_{i,j} = \begin{cases} feel_{i,j} - 1, fit(H_{i,t}) \leq fit(H_{i,t+1}) \\ feel_{i,j} + 1, fit(H_{i,t}) > fit(H_{i,t+1}) \end{cases} \tag{10}$$

$$P_i = \frac{feel_{i,j}}{\sum_{j=i-n}^{i+n} feel_{i,j}} \tag{11}$$

If a neighbor of individual $i$ satisfies $P_1+P_2+\cdots+P_{e-1} < P_{rand} < P_1+P_2+\cdots+P_e$, then the neighbor individual $H_{e,t}$ is selected as the learning object, where $P_{rand}$ is a random number in the interval [0,1].

$$N_e = \max\left(\left\lfloor NP \times \frac{t}{T} \right\rfloor, elitnum\right) \tag{12}$$

Third, individuals will be assigned to different groups according to their fitness. Population $pop$ is divided into two parts of elite individuals $E$ and ordinary individuals $O$, namely $pop = \{E, O\}$. The number of elite individuals $N_e$ is determined by Equation 12, where $t$ represents the current number of iterations, $T$ donates the total number of iterations, and $elitnum$ is the number of initialized elite individuals. Elite individuals have enough resources to guide the evolution of the population, while ordinary individuals can follow the *elite* to update. During the information learning, Equation 13 is used to update the individual.

$$H_{i,t+1} = H_{i,t} + \alpha \times (H_{m,t} - H_{i,t}) + \beta \times (H_{e,t} - H_{i,t}) \quad (13)$$

Where $H_{i,t+1}$ and $H_{i,t}$ represent the position matrix of individual $i$ in the $t$-th and $t+1$th iterations, respectively, $\alpha$ and $\beta$ represent between [0,1] and $\alpha+\beta \leq 1$, $H_{m,t}$ and $H_{e,t}$ represent the best individual and the best neighbor individual in the current iteration, respectively. In general, the first term in Equation 13 is the position of the current individual, the second term enables individuals to learn from the best individuals in the population, and the third term make individuals absorb from outstanding neighbors.

Finally, according to the inertia law, when an individual performs worse in optimization, the individual still has an opportunity to remain in the population. In order to balance exploration and exploitation, the probability of individuals remaining in the group is continuously changing. The probability of retention is calculated by Equation 14, and whether to update its position is determined by Equation 15.

$$P_{accept} = e^{-t} \quad (14)$$

The $t$ in Equation 14 represents the current iteration round. As the number of iterations $t$ increases, the smaller the probability the individual retained. The updated strategy is determined by Equation 15:

$$H_{i,t+1} = \begin{cases} H_{i,t+1}, rand \leq P_{accept} \\ H_{i,t}, rand > P_{accept} \end{cases} \quad (15)$$

In the early stage, individuals have a greater chance to remain. Currently, the model has a powerful ability to explore the solution space. In the later stage, individuals prevent moving to a worse solution space, which can enhance the capability to search locally.

The EPMC model is insufficient converging to good fitness and easy to fall into a locally optimal. Our introduced positional incremental non-linear cosine adaptive crossover and mutation emotional preference migration ensemble clustering (DPNG-EPMC) model can obtain better clustering results by reducing the randomness of optimal solution search and increasing the diversity of individuals in the population.

**Non-linear position weight, directional position increment and Non-linear cosine adaptive crossover and mutation method**

- Non-linear position weight for non-local convergent

As can be seen from Equation 13 in the EPMC model, it only considers the global optimal and excellent neighbors of the current iteration when updating the individual. Therefore, analogous to the particle swarm algorithm, we add the position information of the previous iteration with the inertial weight $w^t$. Therefore, individuals can consider the historical information to help update the position.

In particle swarm optimization, inertia weight $w$ is a linearly decreasing function, which cannot balance the global and local search well. Therefore, we reconstruct the inertia weight into Equation 16, which is the non-linear form. Hence the individual's position update Equation 13 will convert to Equation 18. Besides, the original inertia weight and non-linear inertia weight change trends $w$ in the range of the start value $w_{start}$ and end value $w_{end}$ are shown in Fig. 3.

$$w^t = w_{end} + (w_{start} - w_{end}) \cdot \sin(\frac{\pi}{2}\sqrt{(1-\frac{t}{N})^3}) \quad (1)$$

$$H_{i,t+1} = H_{i,t} + w^t I_{i,t-1} + I_{i,t} \quad (2)$$

$$I_{i,t} = \alpha \times (H_{m,t} - H_{i,t}) + \beta \times (H_{e,t} - H_{i,t}) \quad (3)$$

Where $I_{i,t-1}$ and $I_{i,t}$ represent the position update of individual $i$ in the last iteration and the current iteration, respectively, $\alpha$ and $\beta$ both represent the random numbers between [0,1].

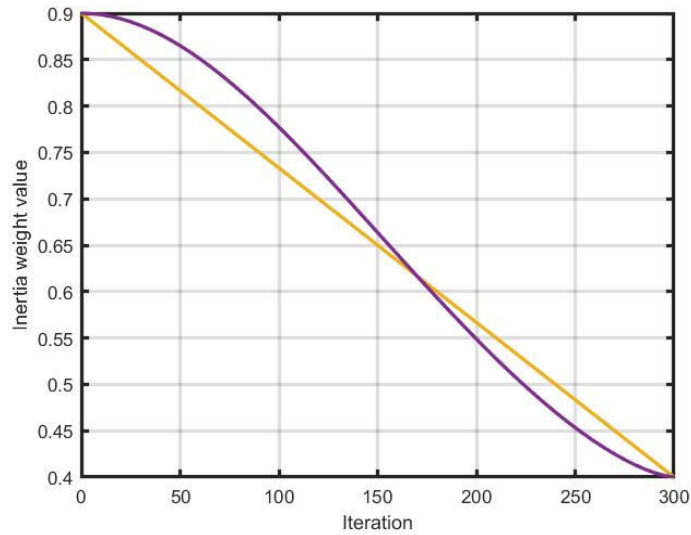

**Figure 4.** Comparison of changes in $w$ under different iteration

As can be seen from **Figure 4**, the yellow line represents the linear inertia weight, and the purple line represents the non-linear sine weight scheme. In the initial stage, the inertia weight $w$ in Equation 16 is larger than the original inertia weight $w$, that is, the second term $w^t I_{i,t-1}$ of Equation 17 obtains a larger value. It allows individuals to search more globally in the early stages without paying too much attention to other individuals nearby (individuals have poor positions in the early stages of iteration). In the second stage, the inertia weight $w$ in Equation 16 is smaller than the original inertia weight $w$, that is, the third item $I_{i,t}$ of Equation 17 gets a larger value. At this time, individuals can search more locally in the later stages, thereby approaching the optimal solution more rapidly.

- Directional position increment for reduce randomness

It can be seen from Equation 13 that individuals are affected by the best individuals in the population and excellent neighbors to update their positions. Although Equation 17 adds the information of historical position with non-linear weights, the search for the optimal solution is still random. As a result, individuals have a broad solution space but a lack of directed search, resulting in unstable results. Therefore, the paper introduces an ingenious searching mechanism based on the position update strategy, as shown in Equation 17. Through auxiliary operation to update the individual, thereby determining the direction of the optimal solution faster and more accurately.

In the searching for the optimal solution, the DPNG-EPMC algorithm adds two-directional position increments $Hr_{i,t}$ and $Hl_{i,t}$ based on Equation 8, which are like two antennae of a beetle, as shown in Equation 19. The beetle antenna search algorithm has been utilized in [40], which has the advantages of fast convergence, easy implementation, and reasonable variable steps to prevent falling into local optimal. We can approximate the behavior of beetles searching for food as the searching for the optimal individual. When the right antennae of the beetle feel the odor concentration of the food is higher than the left side, it means the optimal solution is on the right side of the current position, that is, the fitness value of the right side is smaller. Similarly, when the left antennae of the beetle feel the odor concentration of food is higher than that of the right, it means the optimal solution is on the left of the current position, that is, the fitness value of the left side is smaller.

$$\varphi_{i,t+1} = \sigma^t \cdot I_{i,t} \cdot sign\left(f(Hr_{i,t}) - f(Hl_{i,t})\right) \quad (4)$$

Where $f(Hr_{i,t})$ and $f(Hl_{i,t})$ represent the fitness value of the individual at the positions $Hr_{i,t}$ and $Hl_{i,t}$, respectively, the *sign* is a sign function. When $f(Hr_{i,t}) - f(Hl_{i,t}) < 0$, the function value is -1, and when $f(Hr_{i,t}) - f(Hl_{i,t}) > 0$, the function value is 1. At the same time, when $f(Hr_{i,t}) - f(Hl_{i,t}) = 0$, the function value is 0.

$\sigma^t$ is the step size, which represents the distance the beetle walks in each iteration, as calculated by Equation 20. Besides, *eta* is a constant. In the beginning, the position of the optimal solution is farthest from the beetle, and the step size is suitable to be larger. When the beetle continues to search, the step size gradually turns smaller to ensure precision.

$$\sigma^{t+1} = eta \cdot \sigma^t \quad (20)$$

In addition, the two position vectors $Hr_{i,t}$ 和 $Hl_{i,t}$ can be updated with Equation 21:

$$\begin{cases} Hr_{i,t+1} = Hr_{i,t} + I_{i,t} \cdot d^t/2 \\ Hl_{i,t+1} = Hl_{i,t} - I_{i,t} \cdot d^t/2 \end{cases} \quad (21)$$

Where $d^t$ represents the distance between the two beetles, which can be updated with Equation 22:

$$d^t = \sigma^t/s \qquad (22)$$

Where $\sigma^t$ is the current step size, and $s$ is a non-negative constant.

Therefore, the individual position update formula Equation 17 changed to Equation 23 after adding the position increment,

$$H_{i,t+1} = H_{i,t} + w^t I_{i,t-1} + \eta I_{i,t} + (1-\eta)\varphi_{i,t} \qquad (23)$$

Where $H_{i,t+1}$ and $H_{i,t}$ represent the position matrix of individual $i$ in the $t$-th and $t+1$th iterations, respectively, $\lambda$ represents the increment coefficient, and $\varphi_{i,t}$ represents the direction increment function of the $t$-th iteration. Compared with Equation 17, Equation 23 can search for the optimal solution directionally. Instead of randomly updating the position coordination, thereby helping the model fulfill the clustering task more efficiently.

- Non-linear cosine adaptive crossover and mutation strategy

The individual can learn from the best individual in the population and excellent neighbors. However, still exist some searches that are local optimal rather than global. That is to say, the population falls into a local optimum and does not converge to a true global optimum. Therefore, we propose a novel selection and mutation operation to maintain the population diversity and expect the model to acquire better convergence accuracy and a closer number of clusters.

As we all know, the value of crossover and mutation rate has considerable relevance to the performance of genetic algorithms, which can prevent premature or slow convergence to the optimal. We utilize a novel non-linear cosine adaptive operator to determine individual selection and mutation, as shown in Equation 24 and Equation 25. By non-linearization of the crossover and mutation rate, an individual's fitness is mixed with the average fitness of all individuals. This approach not only preserves the pattern of excellent individuals but also increases the mutation probability of individuals who performed poorly.

$$P_c = \begin{cases} \dfrac{(p_{cmax}-p_{cmin})\cos(\frac{f'-f_{avg}}{f_{max}-f_{avg}}\pi)}{1+\exp(A(\frac{2(f'-f_{avg})}{f_{max}-f_{avg}}))} + p_{cmin}, & f' \leq f_{avg} \\ p_{cmax}, & f' > f_{avg} \end{cases} \qquad (24)$$

$$P_m = \begin{cases} \dfrac{(p_{mmax}-p_{mmin})\cos(\frac{f-f_{avg}}{f_{max}-f_{avg}}\pi)}{1+\exp(A(\frac{2(f-f_{avg})}{f_{max}-f_{avg}}))} + p_{mmin}, & f \leq f_{avg} \\ p_{mmax}, & f > f_{avg} \end{cases} \qquad (5)$$

Here, $p_{cmax}$ and $p_{cmin}$ represent the upper and lower limit crossover rate, respectively. $p_{mmax}$ and $p_{mmin}$ represent the upper limit and lower limit of the mutation rate, respectively.

And $f_{max}$ represents the maximum fitness of individuals in the population, $f_{avg}$ represents the average fitness of individuals in the population. $f'$ represents the smaller fitness of the two individuals involved in the crossover, and $f$ represents the fitness of the variant individual. The cosine function in the denominator of the formula introduces non-linearity, which can prevent the algorithm from premature convergence. At the same time, the molecule is defined by the Sigmoid function frequently utilized in neurons, where A=9.903438, as shown by $\xi(r)$ in Equation 26:

$$\xi(r) = \frac{1}{1+exp(-ar)} \tag{6}$$

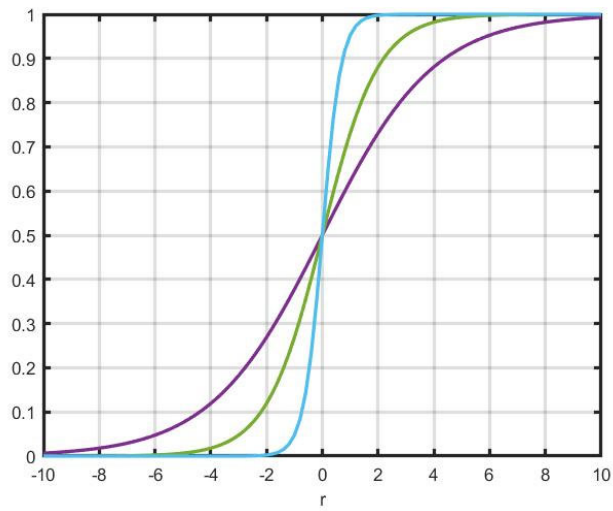

**Figure 5.** Part of the Sigmoid function change trend

**Figure 5** shows the trend of the Sigmoid function. When $ar \geq 9.903438$, $\xi(r)$ is close to 1; when $ar \leq 9.903438$, $\xi(r)$ is close to 0. At this time, the crossover and mutation rate can synchronously adjust with the Sigmoid curve, in which the value is between the maximum fitness and average fitness of all individuals.

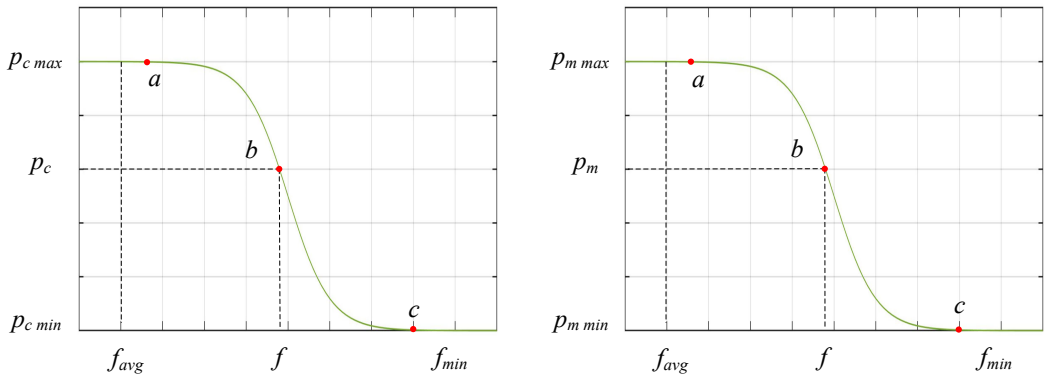

**Figure 6.** Adaptive adjustment curve of crossover rate and mutation rate

In addition, **Figure 6** shows the adaptive adjustment curve of crossover and mutation rate based on the Sigmoid function. When most individuals in the same region or the average fitness are close to the maximum fitness, the individual at this time is relatively poor, so the crossover and mutation rate of most individuals is higher than the value of the linear function, as we can see at point *a* in **Figure 6**. At the same time, at point *c* in **Figure 6**, the algorithm retains the patterns of individuals near the minimum fitness as much as possible. The fitness at point *b* locate at the same position as the linear function, and the crossover and mutation rate have obtained intermediate values. In general, the algorithm can effectively pull apart the individuals whose fitness is in the middle region, thereby promoting the evolution of the whole population and preventing the algorithm from converging to the local optimum.

In general, the schematic diagram and detailed description of the DPNG-EPMC model clustering are shown in **Supplementary Figure S5 and Supplementary Note 5** [51], respectively.

### 2.3 The generative AI SiTime-GAN model for further WWTPs prediction

Like Time-GAN [52] model, SiTime-GAN also consists of four network components: the embedding function, the recovery function, the sequence generator, and the sequence discriminator. It learns to encode features, generate representations, and iterate. The embedding network provides the latent space, and the adversarial network operates within this space, synchronizing the latent dynamics of real and synthetic data through a supervised loss.

Firstly, as a reversible mapping between features and the latent space, embedding and recovery functions should be able to accurately reconstruct the $\tilde{R}$ of the original data $R$. Therefore, the first objective function is the *reconstruction loss*,

$$\mathcal{L}_{Re} = \mathbb{E}_{R\sim p}[\|R - \tilde{R}\|_2] \quad (27)$$

In SiTime-GAN, during training, the generator receives two types of inputs. First, in pure open-loop mode, the generator (autoregressive) takes in embedding $\hat{h}_{R-1}$ to produce the next-generation embedding $\hat{h}_R$. Then, gradients are calculated based on *unsupervised loss*, which is what we expect – either maximizing the *Discriminator* or minimizing the *Generator* to provide the likelihood of correct classification $\hat{y}_R$ for training data $h_R$ and the Generator's output $\hat{h}_R$.

$$\mathcal{L}_U = \mathbb{E}_{R\sim p}[\log y_R] + \mathbb{E}_{R\sim \hat{p}}[\log(1 - \hat{y}_R)] \quad (28)$$

Relying solely on the discriminator's binary adversarial feedback may not be sufficient to motivate the generator to capture the stepwise conditional distribution in the data. To more effectively achieve this, additional loss is introduced for further discipline and learning. In an alternating fashion, it trains in a closed-loop mode where the generator receives the embeddings of real data $\hat{h}_{R-1}$ to generate the next latent vector $\hat{h}_R$. Now, gradients can be computed based on the loss that captures the difference between the distributions $p(h_R)$ and $\hat{p}(\hat{h}_R)$. Applying maximum likelihood results in a familiar supervised loss,

$$\mathcal{L}_S = \mathbb{E}_{R \sim p}\big[\|h_R - g_\mathcal{X}(h_{R-1}, z_R)\|_2\big] \qquad (29)$$

Where $g_\mathcal{X}(h_{R-1}, z_R)$ approximates $\mathbb{E}_{z_R \sim \mathcal{N}}[\hat{p}(h_R, z_R)]$ with one sample $z_R$, which is standard in stochastic gradient descent. In summary, at any step in the training sequence, the model evaluates the difference between the actual next-step latent vector and the synthesized next-step latent vector. $\mathcal{L}_U$ encourages the generator to create authentic sequences, while $\mathcal{L}_S$ further ensures that it produces similar stepwise transitions.

Let $\theta_e, \theta_r, \theta_g, \theta_d$ denote the parameters of the embedding, the recovery, the generator, and the discriminator networks, respectively. The first two components of SiTime-GAN model are trained on both the reconstruction and supervised losses,

$$\min_{\theta_e, \theta_r}(\lambda \mathcal{L}_S + \mathcal{L}_{Re}) \qquad (30)$$

where $\lambda \geq 0$ is a hyperparameter that balances the two losses. Then the generator and discriminator networks are trained adversarially as follows,

$$\min_{\theta_g}(\eta \mathcal{L}_S + \max_{\theta_d} \mathcal{L}_U) \qquad (31)$$

where $\eta \geq 0$ is another hyperparameter that balances the two losses.

The Schematic diagram and detailed description of the SiTime-GAN model are depicted in **Supplementary Figure S6 and Supplementary Note 6 [52-54]**, respectively.

## 3. Results and Discussion

This section briefly introduces the parameters used in the DE-BP neural network prediction model, the proposed DPNG-EPMC clustering algorithm, and the experimental environment. Next, Garson's connection weight is used to analyze the impact of various WWTP feature factors and 1:1 observed-predicted distributions to measure the predictive accuracy.

Lastly, in clustering, the novel DPNG-EPMC algorithm is applied to the clustering analysis of WWTPs under different feature attributes, and we analyze the proportion heatmaps of different microbial communities in various WWTP feature attributes.

### 3.1. Parameter settings and environment

3.1.1. The parameter setting of DE-BP model

The prediction of the DE-BP neural network model is executed in a PyTorch environment, and the experiment's server platform is High-Performance Computing (CREATE HPC) in King's e-Research group. In detail, the King's Computational Research, Engineering and Technology Environment (CREATE) is a tightly integrated ecosystem of research computing infrastructure hosted by King's College London (KCL), which contains 4x NVidia A100 40GB, w/NVLink GPU and 512GB of memory space.

The DE-BP neural network utilized in this topic is structured as 37-8-21, 37-9-51, and 37-15-171 at the Phylum, Class, and Order levels, respectively. The network takes 37 features from WWTPs as inputs and predicts the microbial composition as outputs, categorized into 21, 51, and 171 classes at the Phylum, Class, and Order levels, respectively. **Table 1** contains the parameter settings for the BPNN model and the integration of the DE method.

**Table 1.** Experimental parameter settings of BPNN model and DE method

| Model | Parameter setting |
|---|---|
| BPNN model | The number of epochs *num_epoch* is set to 1500, the learning rate *learn_rate* is set to 1e-2, and the decay of weight *weight_decay is* set to 1e-2. |
| DE method | The crossover probability *p_cross*, the mutation probability *p_mutate, and* the maximum iterations of evolution *MAXGEN* in the DE method are set to 0.4, 0.9, and 20, respectively. |

3.1.2. Initialization and parameter setting of DPNG-EPMC algorithm

To ensure the experiments run smoothly and efficiently, we chose to run the DPNG-EPMC algorithm for clustering on the Lenovo Shenteng 1800 HPCC rack-mounted high-performance optimize server platform, which consists of one control node and eight compute nodes. Each compute node is equipped with two 2.4GHz quad-core processors and 24GB of memory. The operating system running on the server is Red Hat Enterprise Linux 7, and the software computation platform for the experiments is MATLAB R2024b. Table 2 contains the parameter settings for the EPMC algorithm and the newly proposed DPNG-EPMC algorithm.

**Table 2.** Experimental parameter settings of EPMC and newly proposed DPNG-EPMC algorithm

| Algorithm | Parameter setting |
|---|---|
| EPMC | The number of individuals *NP* is set to 15, the number of *elitenum* is set to 2, the number of iterations *T* is set to 100, the initial number of clusters *k* is set to 10, the number of runs is set to 25, set $\alpha = 0.8$ and $\beta = 0.2$, and the max mutation rate *Pmu, max* is set to 0.05. |
| DPNG-EPMC | The number of individuals *NP* is set to 15, the number of elitenum is set to 2, the number of iterations *T* is set to 100, the initial number of clusters *k* is set to 10, the number of runs is set to 25, and set $\alpha = 0.8$, $\beta = 0.2$, sigma=5, d=5, eta=0.8, s=1, MAXGEN=300, $\eta$=0.5, Pcmin=0.5, Pcmax=0.8, Pmmin=0.005, Pmmax=0.05, the max eliminate rate *Pelim* is set to 0.05. |

3.1.3. The parameter setting of generative AI SiTime-GAN model

The Generative AI SiTime-GAN model generates data in a TensorFlow environment. The experimental server platform is also a high-performance computing (HPC) system from King's e-Research group. This platform includes 4 NVidia A100 40GB GPUs with NVLink and 512GB of memory. Table 3 displays the primary parameter settings for the Generative AI SiTime-GAN model.

**Table 3.** Experimental parameter settings of Generative AI SiTime-GAN model

| Algorithm | Parameter setting |
|---|---|

| | | | |
|---|---|---|---|
| Generative AI SiTime-GAN model | The number of samples in each generation *Generative Samples* is set to 1186, the number of iterations *T* is set to 100, the hidden state dimensions *hidden_dim* is set to 24, the number of layers *num_layer* is set to 3, and iterations of the metric computation *metric_iteration* is set to 10. | | |

## 3.2. The prediction results of DE-BP model for WWTPs and analysis

For the prediction of microbial community composition in activated sludge (AS) systems collected from wastewater treatment plants, we conducted a comparison of the predicted microbial community structure and convergence speed between the DE-BP neural network model and the BPNN model across three levels: Phylum, Class, and Order. We used Garson's connection weight to analyze the impact of various WWTP characteristic factors on the predictions, and the predictive accuracy is measured relative to the 1:1 observed-predicted normal distribution.

**The comparison of the predictive performance of DE-BP and BPNN model for microbiome community structure**

In order to compare the two models more intuitively, **Figure 7** shows the experimental results of the BPNN and the DE-BP model on the Phylum, Class, and Order level. As depicted in **Figure 7** (a), the DE-BP model outperforms the BPNN model in terms of mean squared error for OTU proportions in 21 Phylum-level, 51 Class-level, and 171 Order-level predictions across all test datasets. The cross-validation results also show the superiority of the DE-BP model over the BPNN model. **Figure 7** (b-e) shows the error drop curves for one epoch and the comparison between the predicted values and observed values across the test dataset for BPNN and DE-BP models on Phylum, Class, and Order level, respectively. We can observe in **Figure 7** (b-e) that the DE-BP performs a more stable error drop curve and lower predicted error than the BPNN model, and its predicted values are closer to the observed values across each point of the test dataset. More comparison of the predictive performance of DE-BP and BPNN model on other Phylum, Class and Order microbiome community are shown in **Supplementary Figure S7**.

a

| | 21 Phylum-level | 51 Class-level | 171 Order-level |
|---|---|---|---|
| **BPNN model (Train 0-80%; Test 80-100%)** | 9.64658 | 9.16066 | 7.52530 |
| **DE-BP model (Train 0-80%; Test 80-100%)** | 9.05335 | 8.41282 | 7.43444 |
| **BPNN model (Train 20-100%; Test 0-20%)** | 12.9279 | 13.2218 | 9.7340 |
| **DE-BP model (Train 20-100%; Test 0-20%)** | 12.0092 | 11.8987 | 9.5719 |
| **BPNN model (Train 0-60%, 80-100%; Test 60-80%)** | 11.0913 | 11.1636 | 9.3354 |
| **DE-BP model (Train 0-60%, 80-100%; Test 60-80%)** | 10.6645 | 10.1876 | 8.9269 |

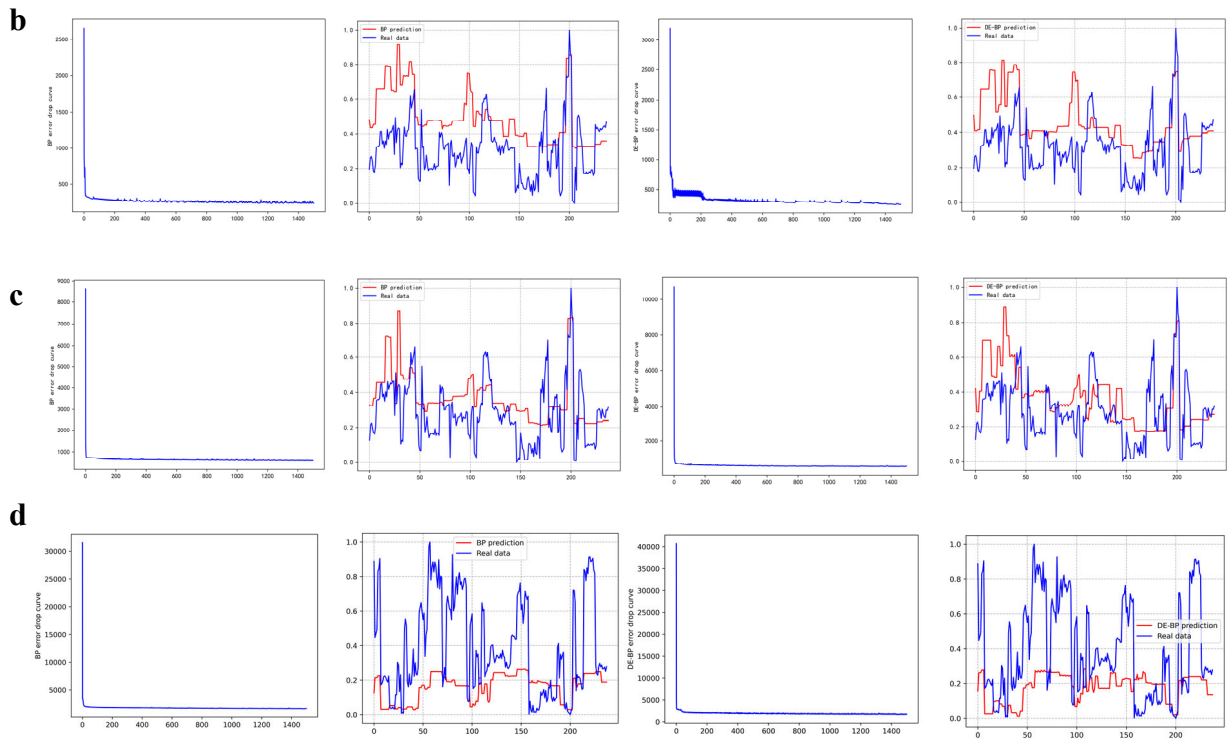

**Figure 7.** The comparison of the predictive performance of DE-BP and BPNN model for microbiome community structure. a, Mean squared error for OTU proportions in 21 Phylum-level, 51 Class-level, and 171 Order-level predictions across all test datasets of BPNN and DE-BP model. b, The error drop curves for one epoch and the comparison between the predicted values and observed values across the test dataset for BPNN and DE-BP models on the Phylum level. c, The error drop curves for one epoch and the comparison between the predicted values and observed values across the test dataset for BPNN and DE-BP models on the Class level. d, The error drop curves for one epoch and the comparison between the predicted values and observed values across the test dataset for BPNN and DE-BP models on the Order level.

**The Garson's connection weight and the 1:1 observed-predicted normal distribution**

**Figure 8** shows the 1:1 observed-predicted normal distribution and Garson's connection weight of the DE-BP and BPNN model for the prediction of the microbiome community structure. Specifically, we can observe from **Figure 8**(a) that the DE-BP model consistently showed an increasing 1:1 observed-predicted values across Order, Class, and Phylum levels, which may be attributed to the increasing complexity of predicting microbial community structures at different levels as the number of microbial community quantities increases, making it more challenging for the model to capture information from the same quantity of WWTP input features. In addition, under different levels, the Garson's connection weights reveal various values **Figure 8** (b). As we can see in **Figure 8** (b), each input feature serves a crucial role in predicting the microbial community structure of the WWTPs and uniformly influences the operation of the whole biochemical system.

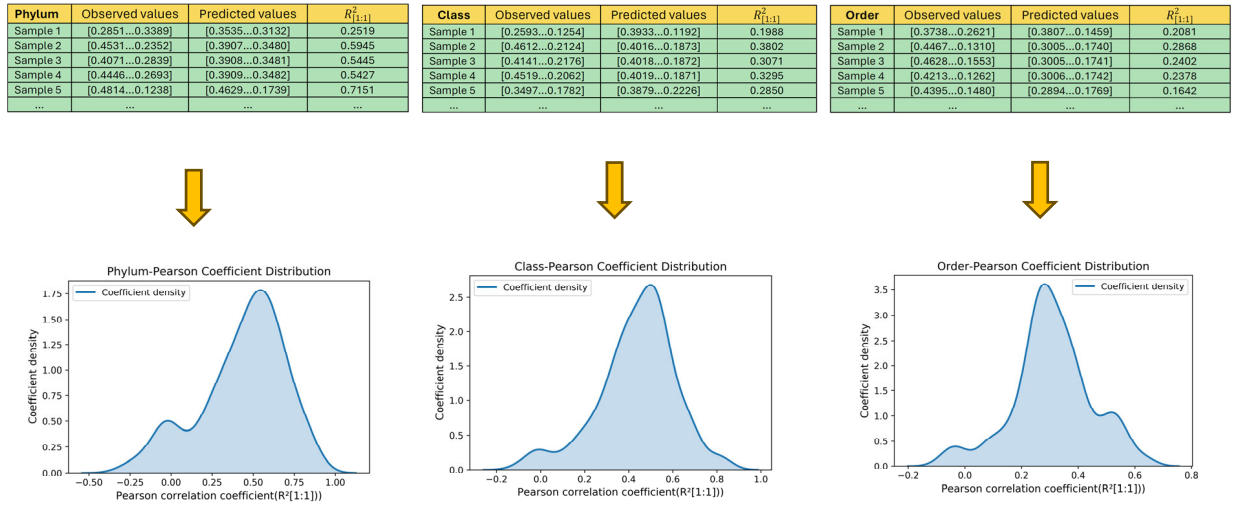

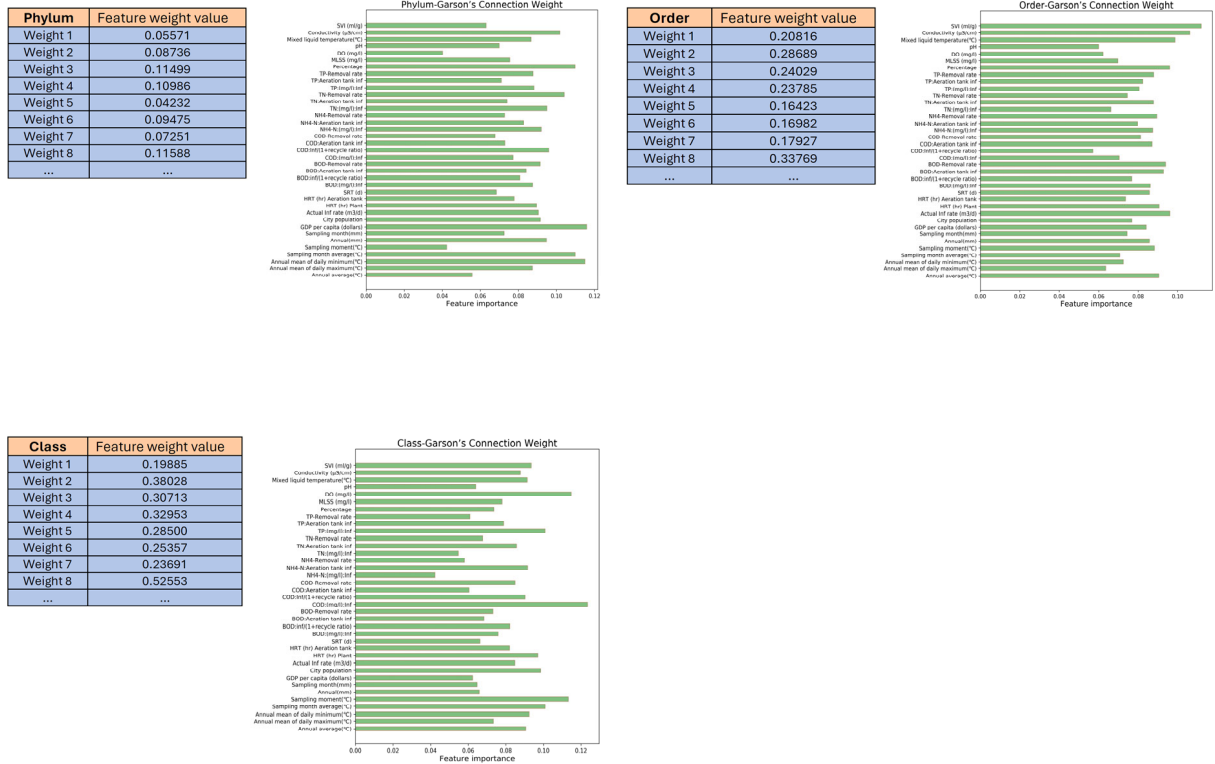

**Figure 8.** The Garson's connection weight and the 1:1 observed-predicted normal distribution of DE-BP and BPNN model for the microbiome community structure prediction. a, The 1:1 observed-predicted normal distribution of DE-BP model on Phylum, Class, and Order level. b, The Garson's connection weight of various features across DE-BP model on Phylum, Class, and Order level.

### 3.3 The clustering results of DPNG-EPMC for WWTPs under different characteristic attributes

For the clustering analysis of WWTPs under various feature attributes: the influent features, the effluent features, the reactor operational features, and the environmental and geographical features, we utilize the novel DPNG-EPMC algorithm to properly clustering global WWTPs to get the main microbial community composition on the Phylum, Class, and Order levels of different WWTPs. We generate heatmaps to depict the proportion of different microbial communities among these WWTPs. As we can see in **Figure 9** (a), different colors represent different feature attributes, and the DPNG-EPMC algorithm utilized these 37 features from WWTPs and the microbial community composition data at three levels to perform the clustering analysis. **Figure 9** (b-e) shows the heatmaps of the 'core' microbial community structure in WWTPs at the Phylum, Class, and Order levels under four feature attributes: the environmental and geographical features, the influent features, the effluent features, and the reactor operation features. Specifically, we can observe from **Figure 9** (b-e) that in different clusters of WWTPs, the microbial composition at various taxonomic levels is remarkably similar. However, despite this, there are significant differences in the microbial community compositions among various WWTP clusters. For instance, in the 'core' microbial community structure under the effluent features at the Class level in WWTPs, the most abundant OTUs, the 'Gammaproteobacteria' and the 'Bacteroidia', exhibit opposite distributions across different plants. In addition, in the 'core' microbial community structure under the effluent features at the Phylum level in WWTPs, the OTU 'Cyanobacteria' in 'Cluster5' shows a significant and anomalous abundance, which may potentially be attributed to the substantial and random impact of effluent features on the microbial distribution in the water treatment plant, resulting in an unstable microbial community distribution.

a

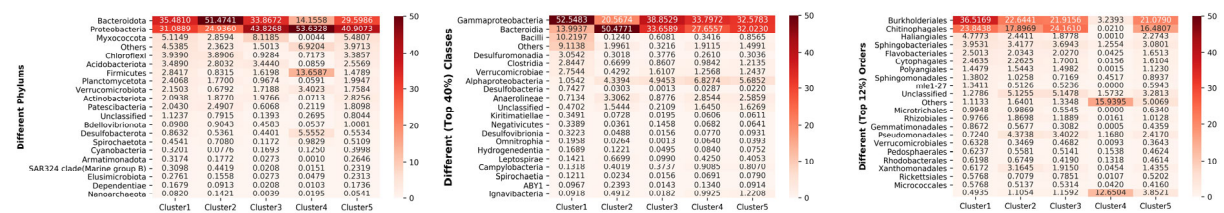

b

c

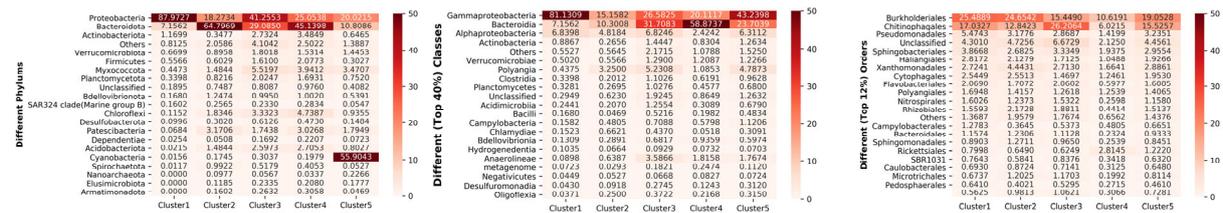

d

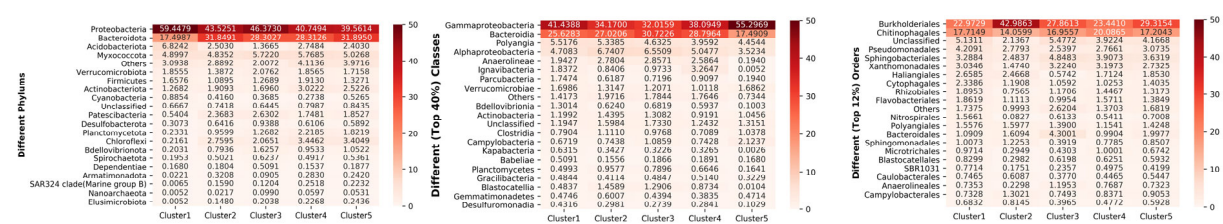

e

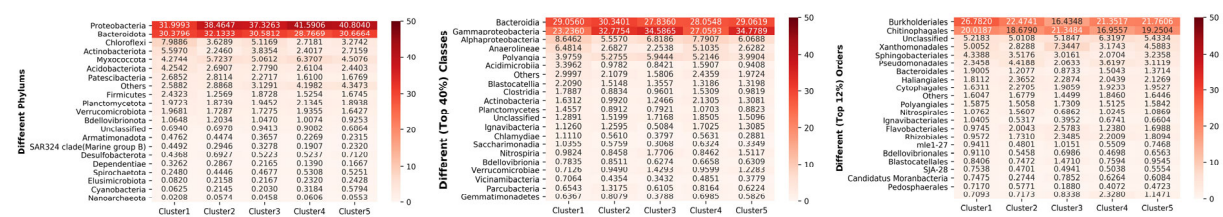

**Figure 9.** The heatmaps of varying microbial community proportions in WWTPs under different feature attributes. a, The features at the Phylum, Class, and Order levels under four feature attributes: the environmental and geographical features, the influent features, the effluent features, and the reactor operation features. b, The heatmaps of 'core' microbial community structure in WWTPs under the influent features at Phylum, Class, and Order levels. c, The heatmaps of 'core' microbial community structure in WWTPs under the effluent features at Phylum, Class, and Order levels. d, The heatmaps of 'core' microbial community structure in WWTPs under the reactor operation features at Phylum, Class, and Order levels. e, The heatmaps of 'core' microbial community structure in WWTPs under the environmental and geographical features at Phylum, Class, and Order levels.

**The centroids and species feature under different clusters after PCA dimension reduction heatmap at Class level on effluent feature attribute**

**Figure 10** shows the heatmap of centroids and the line chart of species features under different clusters after PCA dimension reduction at a Class level under the effluent feature attribute. As we can observe from **Figure 10** (a), different colors and depths represent their values after PCA dimensionality reduction. After the PCA dimension reduction, different clusters show completely different feature distributions. That is probably due to the fact that the clustering centroids are relatively far away, i.e., the different classes of WWTPs have been sufficiently differentiated based on their OTU features. At the same time, **Figure 10** (b) shows the line chart of all the WWTPs species features under each cluster after PCA dimension reduction at the Class level under the effluent feature attribute. It can be clearly seen that the first cluster has the highest number of water plants, which indicates that most of the water plants belong to the same class of bacterial distribution. In contrast, several others have a small number of specific distributions. After PCA dimension reduction, these two plots also visualize the global distribution of WWTPs while creating far-reaching implications for ecosystem bacterial health

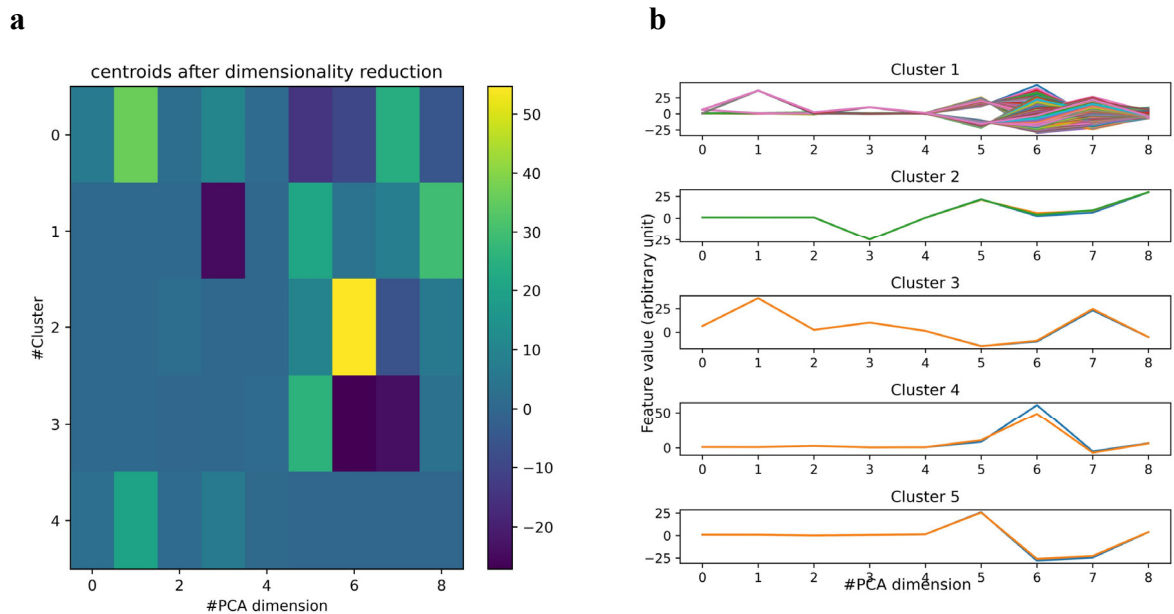

**Figure 10.** The heatmap of centroids and the line chart of species features under different clusters after PCA dimension reduction at Class level under effluent feature attribute. a, The heatmap of centroids after PCA dimension reduction at Class level under effluent feature attribute. b, The line chart of species feature under different cluster after PCA dimension reduction at Class level under effluent feature attribute.

**The effluent features heatmap and average effluent features line chart under different clusters at Class level**

The heatmap of effluent features and line chart of average effluent features under different clusters at the Class level are shown in **Figure 11**. It can be inferred from **Figure 11** (a) that the feature 'BOD-removal rate', 'NH4-removal rate' and 'TP-removal rate' always have similar values under different clusters. Then, the feature 'COD-removal rate' also indicates that the wastewater of the relevant WWTPs is more contaminated by organic matter. In addition, the high value of the feature 'TN-removal rate' also indicates that the area of interest includes more frequent agricultural activities (e.g., heavy use of fertilizers). For **Figure 11** (b), the five effluent feature attributes have almost the same trend of mean values on different clusters, except for cluster 4 and cluster 5, which show rare identical values on feature 'NH4-removal rate', which may be due to the close geographical proximity of the water plants at the center of the clusters concerned.

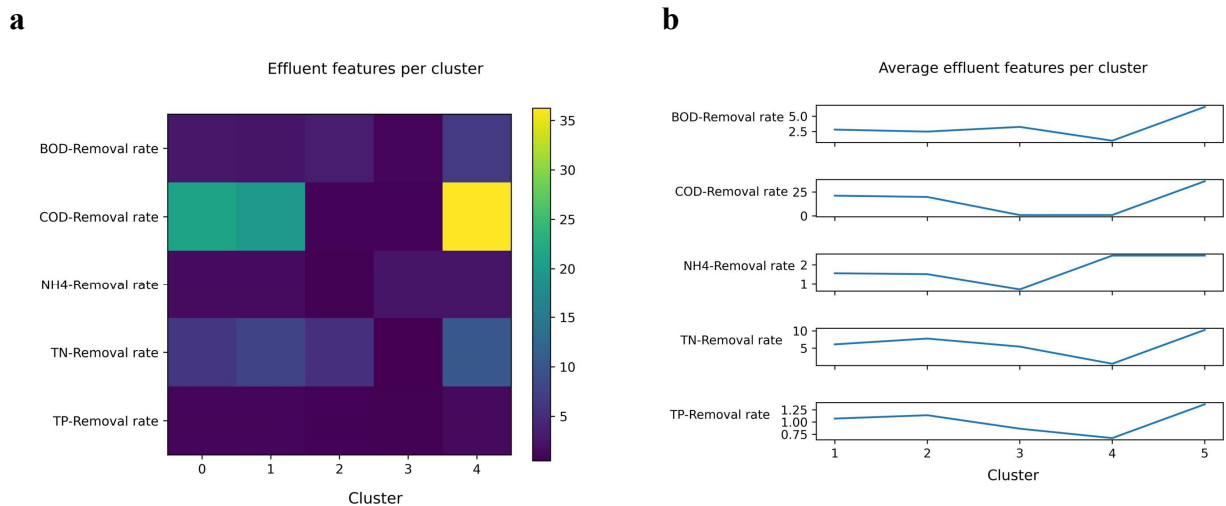

**Figure 11.** The heatmap of effluent features and the line chart of average effluent features under different clusters and at Class level. a, The effluent features heatmap under different clusters at Class level. b, The average effluent features line chart under different clusters at Class level.

**The OTU proportion on a log scale heatmap and average OTU proportion on a log scale line chart under different clusters at Class level**

The heatmap of OTU proportion on a log scale and line chart of average OTU proportion on a log scale under different clusters at the Class level is shown in **Figure 12**. As we can derive from **Figure 12** (a), except for the cluster centroid in the first column, the other cluster centroids have almost similar distributional on the log scale. That echoes the results of the line chart in **Figure 10** (b), where most of the water plant bacterial distributions belong to the same class, and reflects that human wastewater from different continents around the globe contains the same OTU composition. **Figure 12** (b) shows the average OTU proportion on a log scale line chart under different clusters at the 51-Class level, which also echoes the results of **Figure 12** (a) that the consistency of the distribution is concentrated on the last four clusters centroids. The combination of the Effluent water plant features in **Figure 11** and the distribution of OTU features on the Class level in **Figure 12** is the feature representation in **Figure 10**, except that the PCA operation in **Figure 10** downscaled and compressed all the features.

a

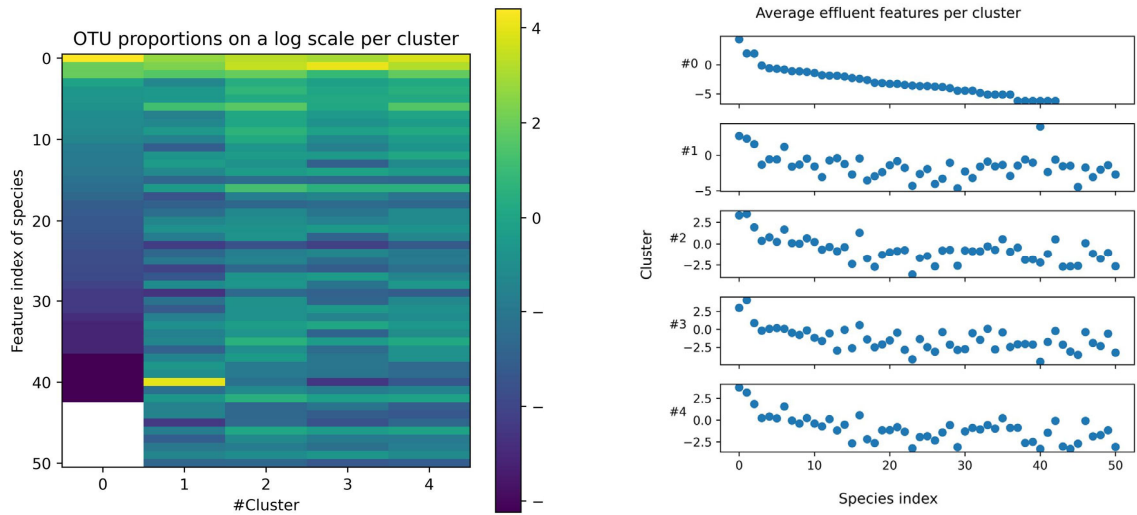

**b**

**Figure 12.** The heatmap of OTU proportion on a log scale and the line chart of average OTU proportion on a log scale under different clusters at Class level. a, The OTU proportion on a log scale heatmap under different clusters at Class level. b, The average OTU proportion on a log scale line chart under different clusters at Class level.

**The first three-dimension species features under different clusters after PCA dimension reduction and heatmap of mutual information between effluent and species features**

**Figure 13** shows the first three-dimensional species features under different clusters after PCA dimension reduction and heatmap of mutual information between effluent and species features. It can be deduced from **Figure 13** (a) that different colors represent different clusters, visualizing a portion of the feature distribution after the PCA dimensionality reduction. Besides, **Figure 13** (b) depicted the mutual information between effluent and species features, where different colors represent different amounts of OTU information included in different effluent feature attributes, and the colors of the five vertical axis labels also distinguish different effluent feature attributes. Meanwhile, the five different colors of OTU names "Gammaproteobacteria", "Bacteroidia", "Alphaproteobacteria", "Phycisphaerae", and "Campylobacteria" in the horizontal axis correspond sequentially to the effluent feature attributes "TN-Removal rate", "BOD-Removal rate", "TP--Removal rate", "NH4-Removal rate" and "COD--Removal rate" of the highest information content, respectively. This figure also shows the correlation distribution of different effluent feature attributes with different OTUs at the class level.

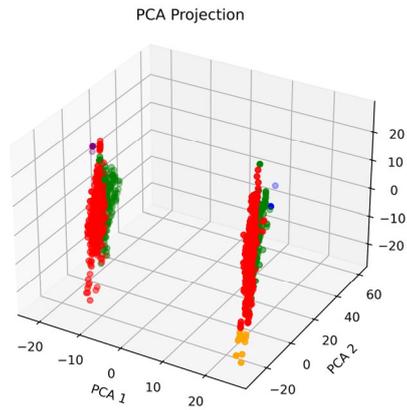 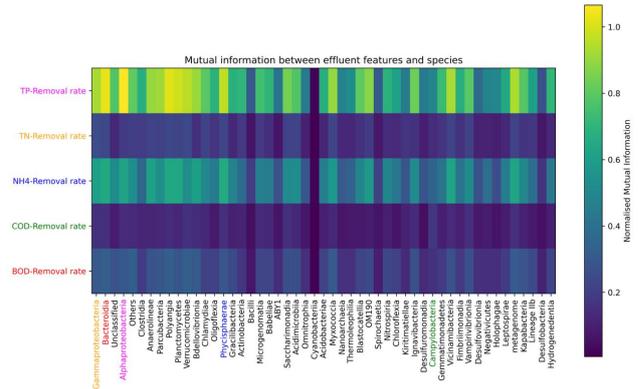

**Figure 13.** The first three-dimension species features under different clusters after PCA dimension reduction heatmap of mutual information between effluent and species features. a, The first three-dimension species features under different clusters after PCA dimension reduction. b, The heatmap of mutual information between effluent and species features.

## 3.4. The predictions of generate trained BPNN and DE-BP model for WWTPs and analysis

For the data generation of different feature attributes and the microbial community structure, we employ a model similar to Time-GAN to generate data for the further microbial community structure prediction of WWTPs at the Phylum, Class, and Order levels. The experiment was repeated five times for each trial, and the average mean squared error for OTU proportions in 21 Phylum-level, 51 Class-level, and 171 Order-level predictions across all test datasets is shown in **Figure 14** (a). Then, we compared the predicted microbial community structure and convergence speed between the DE-BP neural network model and the BPNN model on the original data and merged new data across three levels: Phylum, Class, and Order. In addition, we compared the prediction results distribution of the DE-BP algorithm and the BPNN model for the microbial community structure over five trials of repeated experiments on the original data as well as on the merged new data at the Phylum, Class, and Order levels. Finally, the predictive performance and heatmap of the DE-BP and BPNN models were compared, and they were trained with incremental training sets after merging 100 or 948 generated new data points (948 is the same scale as the training data).

**The comparison of the predictive performance of DE-BP and BPNN model on the original data and merged new data**

**Figure 14** (b-e) shows the error drop curves for one epoch and the comparison between the predicted values and observed values across the test dataset for BPNN and DE-BP models on Phylum, Class, and Order level across original data and merged new data, respectively. As shown in **Figure 14** (b-e), the DE-BP algorithm performs a more stable error drop curve than the BPNN model across original and merged new data. The predicted and observed values of both BPNN and DE-BP models are closer to the merged new data than the original data.

**a**

|  | 21 Phylum-level | 51 Class-level | 171 Order-level |
|---|---|---|---|
| **BPNN model (Original data)** | 9.64658 | 9.16066 | 7.52530 |
| **DE-BP model (Original data)** | 9.05335 | 8.41282 | 7.43445 |
| **BPNN model (Merged data)** | 9.40432 | 9.04253 | 7.53677 |
| **DE-BP model (Merged data)** | 8.74552 | 8.08083 | 7.21274 |

**b**

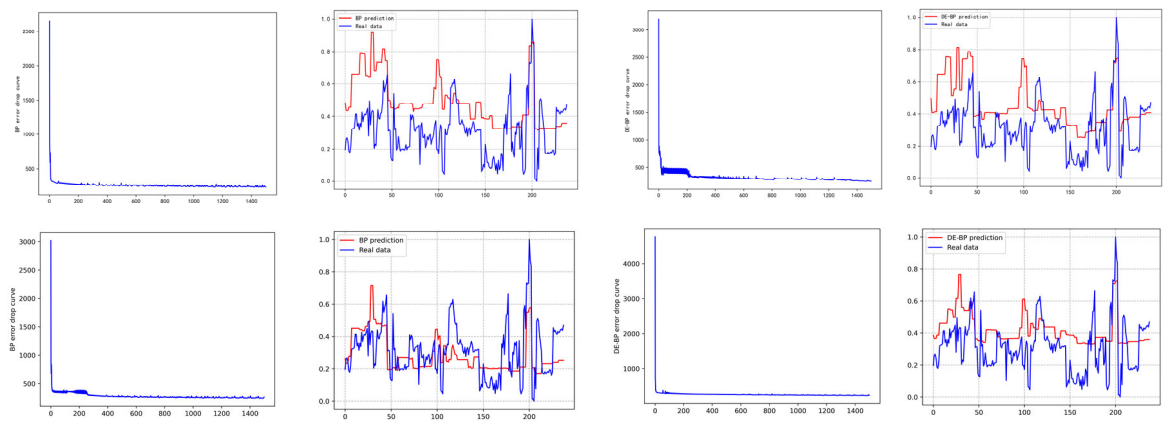

**c**

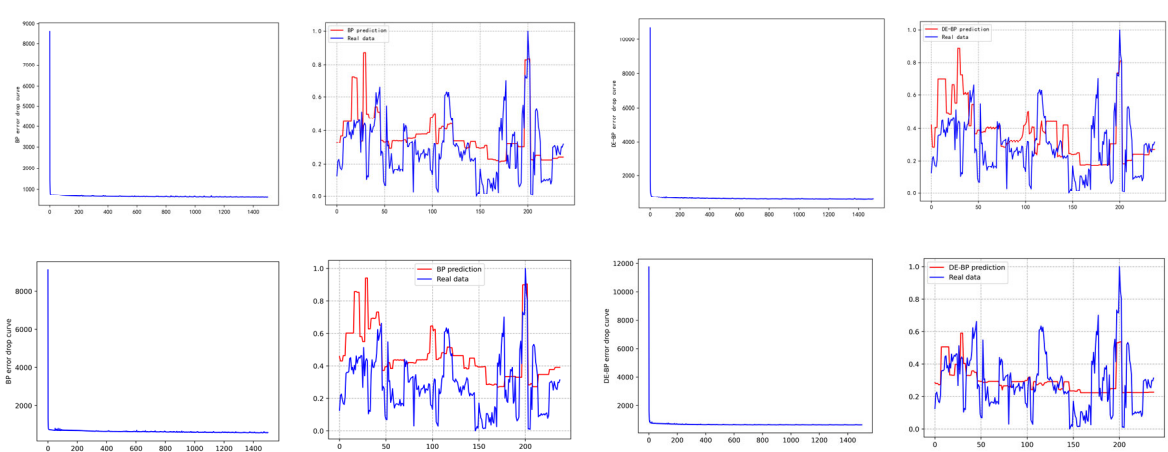

**d**

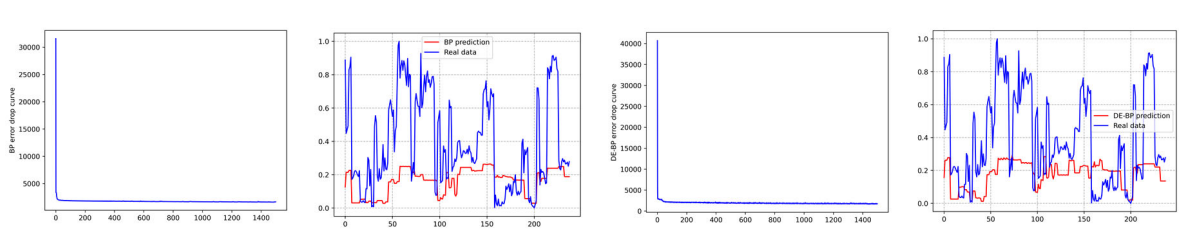

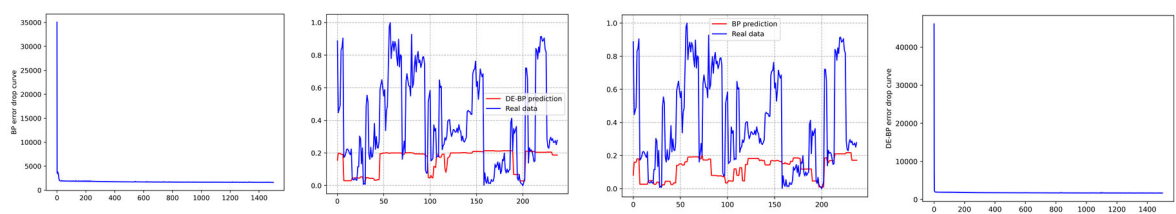

**Figure 14.** The comparison of the predictive performance of DE-BP and BPNN model on the original data and merged new data. a, Mean squared error for OTU proportions in 21 Phylum-level, 51 Class-level, and 171 Order-level predictions on the original data and merged new data BPNN and DE-BP model. b, The error drop curves for one epoch and the **comparison** between the predicted values and observed values across the original data and merged new data for BPNN and DE-BP models on the Phylum level. c, The error drop curves for one epoch and the comparison between the predicted values and observed values across the original data and merged new data for BPNN and DE-BP models on the Class level. d, The error drop curves for one epoch and the comparison between the predicted values and observed values across the original data and merged new data for BPNN and DE-BP models on the Order level.

**The prediction results distribution of the DE-BP and BPNN model over five trials repeated experiments on the original data and merged new data**

To further analyze the effects of data augmentation, **Figure 15** (a-c) shows prediction results distribution of DE-BP and BPNN models over five trials repeated experiments on the original data and the merged new data at the Phylum, Class, and Order levels. As can be seen in **Figure 15** (a-c), the error distribution of the prediction results from the DE-BP model consistently lies below that of the BPNN model, as well as the error distribution of the prediction results from merged new data consistently lies below the original data. In other words, the DE-BP and the BPNN models trained on augmented data achieved superior predictive performance compared to their respective models trained on the original data.

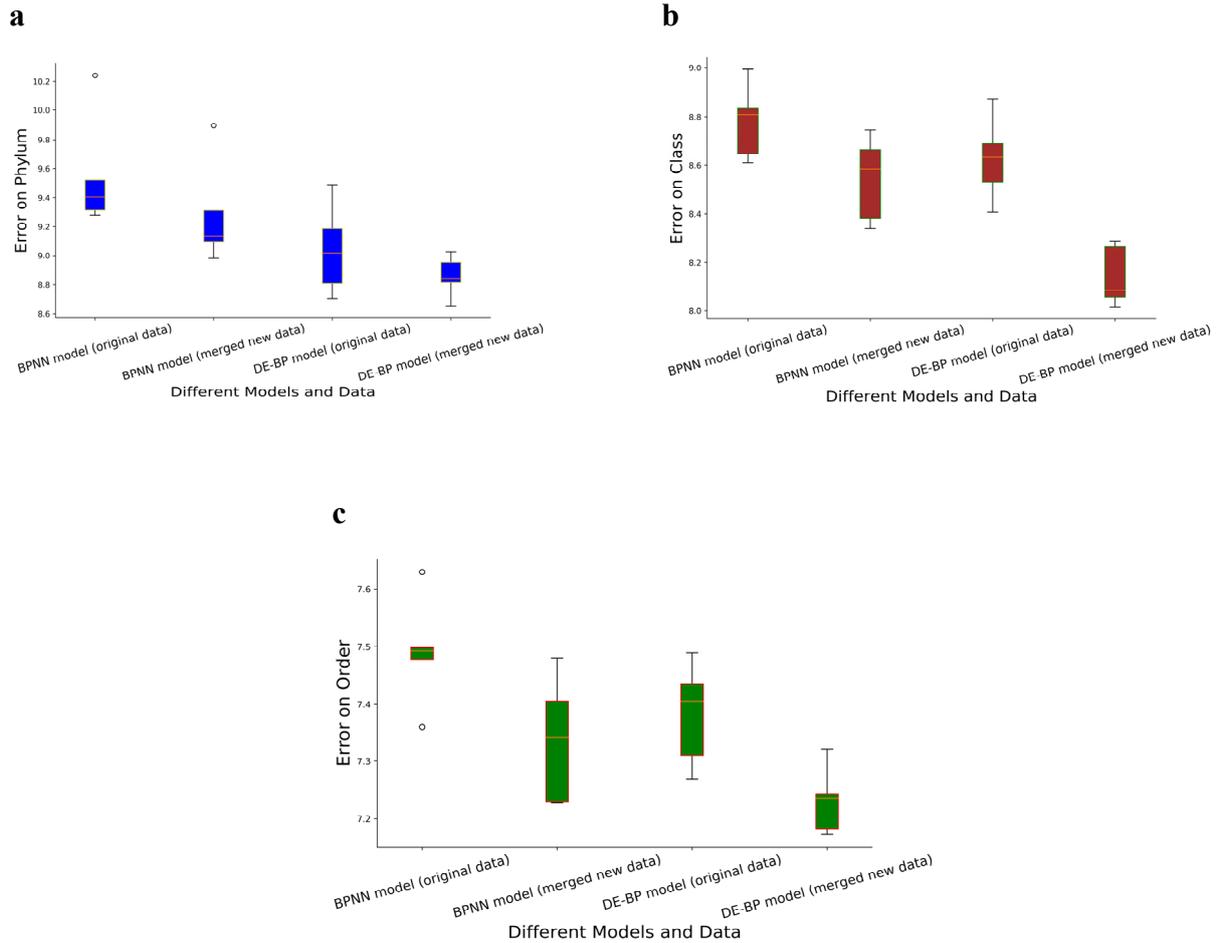

**Figure 15.** The comparison of the predictive results distribution over five trials repeated experiments on the original data and merged new data. a, The prediction results distribution of DE-BP and BPNN model on the original data and merged new data at the Phylum level. b, The prediction results distribution of DE-BP and BPNN model on the original data and merged new data at the Class level. c, The prediction results distribution of DE-BP and BPNN model on the original data and merged new data at the Order level.

**The predictive performance of the DE-BP and BPNN models after merging 100 or 948 generated new data points at the Phylum-level**

To test the quality of the generated data, "Perfect" generated data would match the performance of the original data. We compare the predictive performance of the DE-B P and BPNN models after merging 100 or 948 generated new data points on the Phylum level. **Figure 16** (a-c) depicts the predictive performance comparison of the DE-BP and BPNN models after merging different sizes of generated new data points. Specifically, **Figure 16** (a) shows the Observation-Prediction error for OTU proportions in Phylum-level across all test datasets trained on original data, merged 100-generated new data, and merged 948-generated new data, respectively. Then, **Figure 16** (b-c) shows the error drop curves for one epoch and the comparison between the predicted values and observed values for BPNN and DE-BP models on the Phylum level after merging 100 or 948 generated new data points. As we can see in **Figure 16** (a), the DE-BP algorithm outperforms the BPNN model in terms of error on both the original dataset and the merged generated new dataset. We expect the mean error to drop on the test dataset when more generated data is added to the test dataset. In contrast, as we can see in **Figure 16** (a-c), adding more "bad" generated data to training the model would degrade the performance of the neural network on the test dataset, as it would serve only to confuse the neural network, and cannot provide additional meaningful information.

a

|  | Original data | 100-Generated new data | 948-Generated new data |
|---|---|---|---|
| **BPNN model** | 9.55106 | 9.40432 | 17.1642 |
| **DE-BP model** | 9.04034 | 8.74552 | 10.2398 |

b

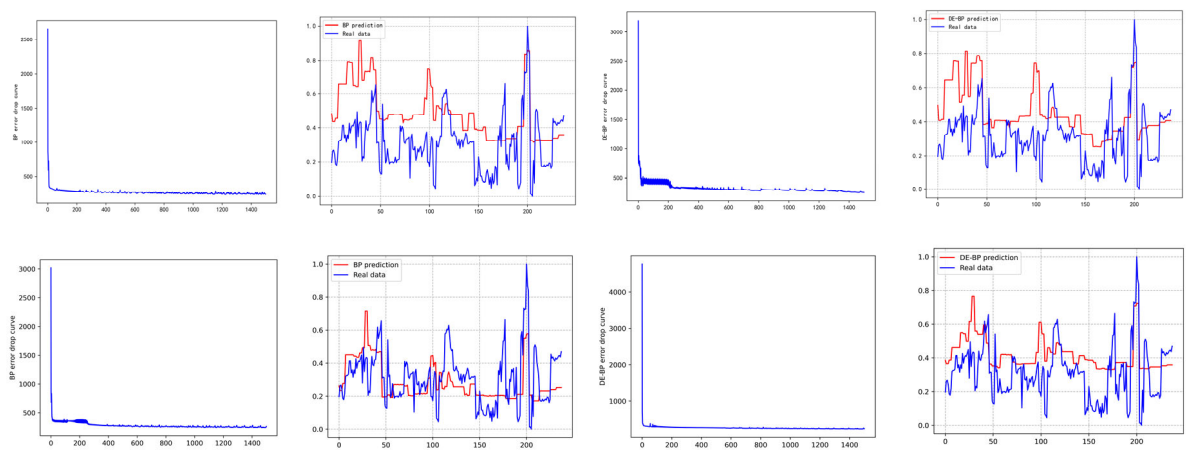

c

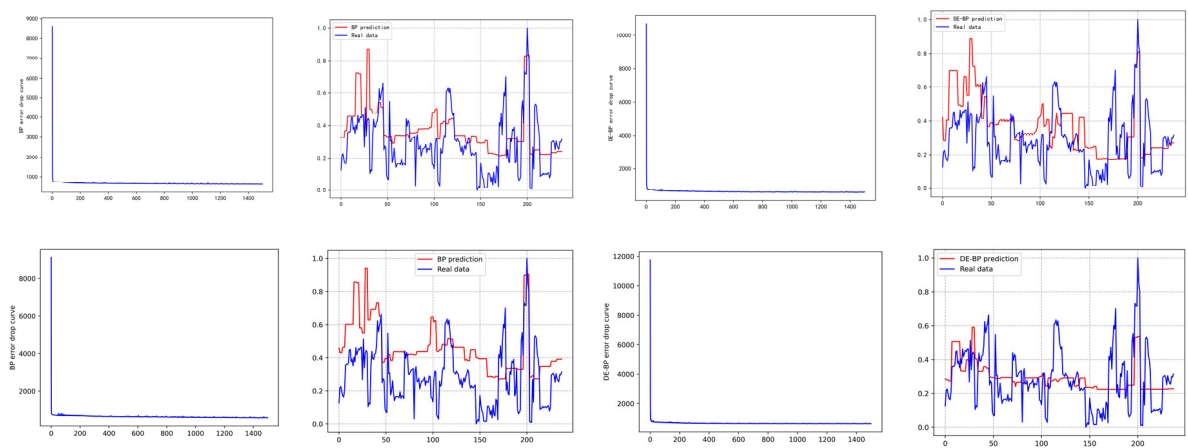

**Figure 16.** The predictive performance of the DE-BP and BPNN models after merging 100 or 948 generated new data points. a, The predictive performance of the DE-BP and BPNN model after merging 100 or 948 generated new data points. b, The error drop curves for one epoch and the comparison between the predicted values and observed values for BPNN and DE-BP models on the Phylum level across the 100 generated new data points. c, The error drop curves for one epoch and the comparison between the predicted values and observed values for BPNN and DE-BP models on the Phylum level across the 948 generated new data points.

**The comparisons of the heatmaps on the Observations data, Predictions based on original data, and Predictions based on Original-Generated data of the DE-BP model**

As we can see, **Figure 17** (a-c) shows the heatmaps of the Observations data, Predictions of the DE-BP model based on original data, and Predictions of the DE-BP model based on Original and generated 100 data at the Phylum, Class, and Order level. In detail, the predicted distribution of the DE-BP model on the original data and the Original-Generated 100 data is roughly similar. However, the predicted value on the Original-Generated 100 data is more distributed in the greater-than-median portion compared to the original data, which may be due to the generated data being closer-to-median, making the training of the DE-BP model more stable and convergent. Besides, the predictions on Original-Generated 100 data are smoother than on original data at the Phylum, Class, and Order level, which may result in the dataset not being large enough to fully exploit the neural network's capacity. Therefore, more data points need to be merged to enhance the neural network's generalization.

a

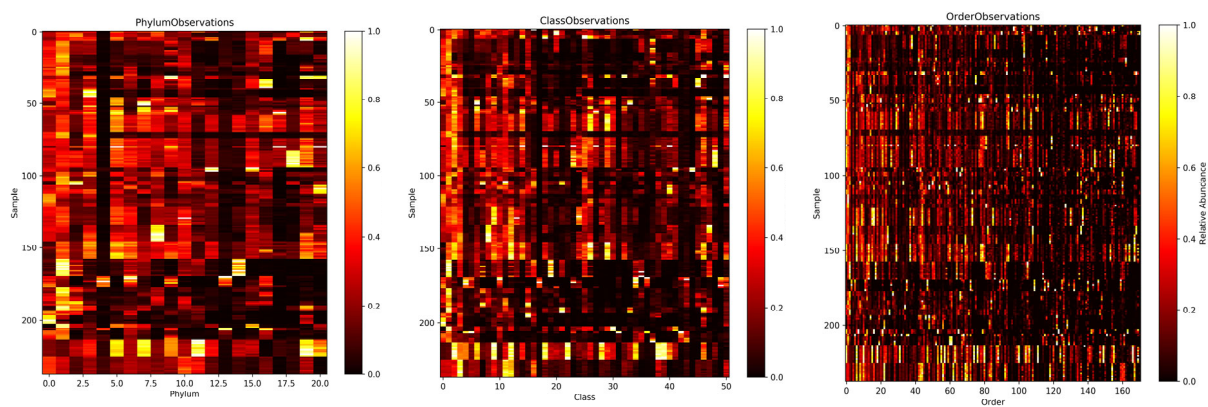

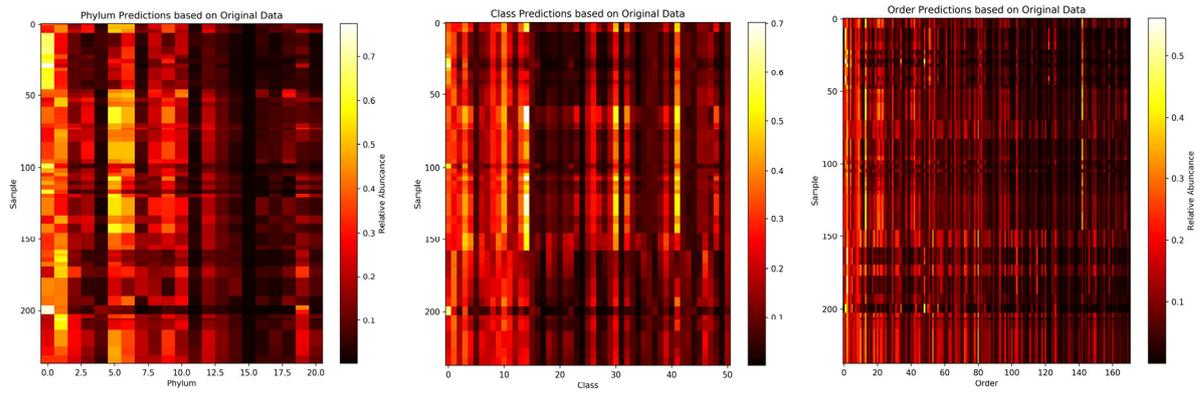

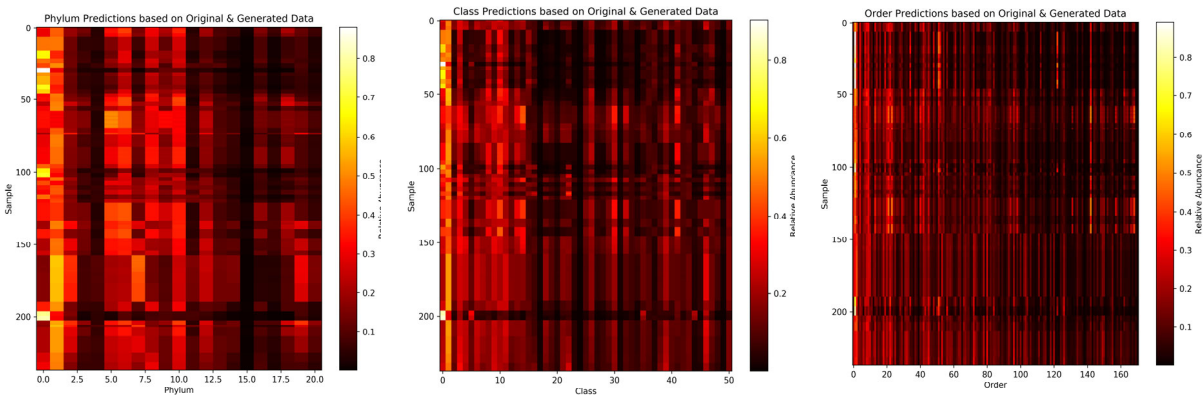

**Figure 17.** The comparisons of the heatmaps of the Observations data, Predictions based on original data, and Predictions based on Original- Generated data on the DE-BP model at the Phylum, Class, and Order level. a, The comparisons of the heatmaps of the Observations data on the DE-BP model at the Phylum, Class, and Order level. b, The comparisons of the heatmaps of the Predictions based on Original data on the DE-BP model at the Phylum, Class, and Order level. c, The comparisons of the heatmaps of the Predictions based on Original-Generated on the DE-BP model at the Phylum, Class, and Order level.

**The predictive performance and heatmaps after merging 100 or 948 generated new data points of the DE-BP model at the Phylum-level**

For the comparisons of predictive performance after merging 100 or 948 generated new data points of the DE-BP model, we utilized the RMSE bar chart and heatmaps on original data, Generated 100 data, and Generated 948 data at the Phylum level to analyze. As shown in **Figure 18** (a), "perfect" added 100 generated data matches the performance of the original data on the DE-BP model. However, adding an extra 948 generated data would degrade the performance of the DE-BP model, as it would confuse the DE-BP model. That means adding some generated data improved the performance, but too much degraded the performance. Besides, the predicted heatmap of the DE-BP model on original data, which Generated 100 data and 948 data at the Phylum level, is depicted in **Figure 18** (b-d). As shown in **Figure 18** (b-d), the predicted value on the Generated 948 data is more distributed in the close-to-minimum portion compared to the Generated 100 data and original data. As the scale of the merged data increases, the predictions become smoother. The DE-BP model training has been affected by the large amount of new data and has lost more detailed information from the original dataset.

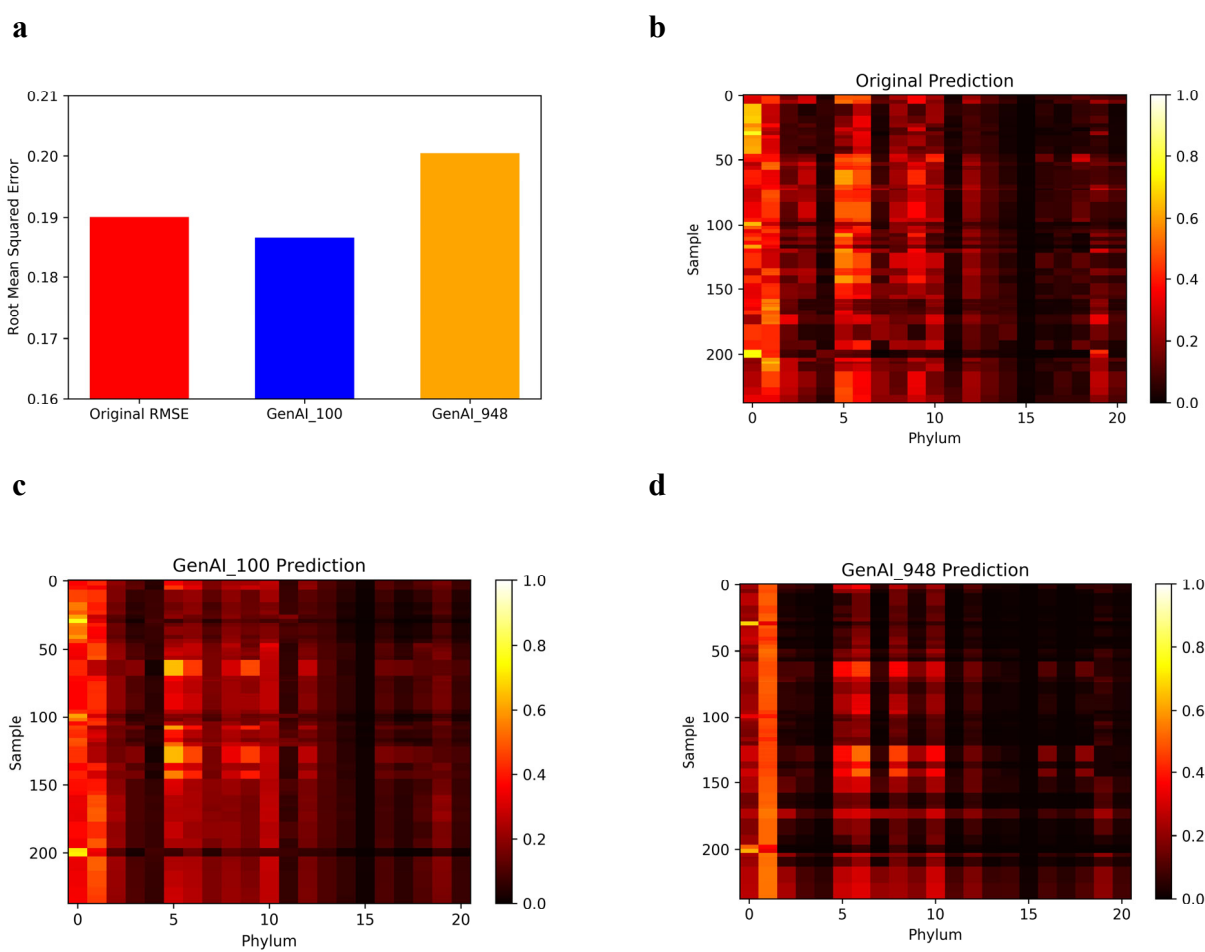

**Figure 18.** The predictive performance and the heatmaps after merging 100 or 948 generated new data points of the DE-BP model at the Phylum-level. a, The predictive performance after merging 100 or 948 generated new data points of the DE-BP model at the Phylum-level. b, The predicted heatmaps of original data points of DE-BP model on the Phylum level. c, The predicted heatmaps after merging 100 generated new data points of DE-BP model on the Phylum level. d, The predicted heatmaps after merging 948 generated new data points of DE-BP model on the Phylum level.

## 4. Conclusion

On the basis of the DE-BP model, the DPNG-EPMC algorithm, and the SiTime-GAN model, the paper performed the prediction of microbial community structure in WWTPs, clustering WWTPs under various feature attributes, and generating substantial data on different attributes of WWTPs to train the DE-BP model further, respectively. Firstly, the paper utilized the DE-BP neural network with data on various feature attributes to predict the microbial community structure of AS systems from global WWTPs. Secondly, the paper proposed the DPNG-EPMC algorithm, which enables the clustering analysis of similar WWTPs based on different feature attributes. Lastly, the SiTime-GAN model generates feature data for different WWTP attributes of WWTPs and employs the newly generated data to train the DE-BP model further.

Then, the theoretical analysis and process derivation of the DE-BP model, the DPNG-EPMC algorithm, and the SiTime-GAN model are given. Numerous cross-validation experiments were performed to compare the DE-BP model with the BPNN models on the original data. New data on the Phylum, Class, and Order levels were generated and merged. Besides, the Garson's connection weight and the 1:1 observed-predicted normal distribution is adopted to compare the DE-BP and BPNN model; the heatmaps is applied to identify the 'core' microbial community structure in WWTPs under different feature attributes; the heatmap and line chart are utilized to analysis the centroids and species feature under different clusters after PCA dimension. Finally, we also compare predictive performance and heatmaps after merging 100 or 948 generated new data points with the original data on the DE-BP model, and the prediction results distribution of the DE-BP and BPNN models over five trial repeated experiments on the original data and merged new data.

However, related methods still have many drawbacks. For instance, the improvement of the prediction accuracy of the BPNN network with the addition of differential evolution for microbial community is still relatively small, and the data generated based on the SiTime-GAN augmentation is still relatively smooth, i.e., failing to mimic the characteristics of the original data fully. In the future, we will focus on propose new neural network to predict the interactions between the microbiome and the environment as well as the composition of the microbiome more concise; applied superior optimization algorithm based on emotional preference and migration behavior to classify microbial features and identify the 'core' microbial community to assess the stability and health of ecosystems; propose newly generated model to understand the spatial and temporal characteristics of the microbiome to optimize biotechnological applications further, improved control of microbiome communities.

# Declarations

**Author Contributions**

Mingzhi Dai was responsible for the paper's framework establishment, algorithm implementation, code execution, communication organization, and paper writing; acknowledges for Weiwei Cai for the paper's overall data collection, WWTP expertise review, supplementary experiments, and idea exchange; acknowledges for Thomas Vinestock for the paper's prediction algorithm, generative algorithm content review, supplementary experiments, code execution, and idea exchange; acknowledges for Lennart Otte for the paper's clustering algorithm content review, supplementary experiments, code execution, and idea exchange; acknowledges for Xiang Feng for the paper's format suggestions, content suggestions; acknowledges for Huiqun Yu for the paper's content suggestions; acknowledges for Weibin Guo for the paper's experimental suggestions; and acknowledges for Miao Guo for the paper's overall framework design, content integration, background synthesis, team organization, experiment determination, and idea exchange.

**Competing Interests**

The authors declare that there is no commercial or associative interest that represents a conflict of interest in connection with the work submitted, there is no professional or other personal interest of any nature or kind in any product, service, and/or company that could be construed as influencing the position presented in.

**Data Availability Statement**

The part of running code, as well as data and other materials, are available in the following GitHub folder: https://github.com/MGuo-Lab/Prediction-Clustering-Generation-in-WWTP/tree/main. Correspondence and requests for further materials are available from the corresponding author, upon reasonable request (miao.guo@kcl.ac.uk).

# Supporting Information

**Prediction, Generation of WWTPs microbiome community structures and Clustering of WWTPs various feature attributes using DE-BP neural network model, SiTime-GAN model and DPNG-EPMC ensemble clustering algorithm with modulation of microbial ecosystem health**

*Mingzhi Dai, Weiwei Cai, Xiang Feng, Huiqun Yu, Weibin Guo and Miao Guo\**

**Supplementary Figure S1: Schematic diagram of the BP model framework**

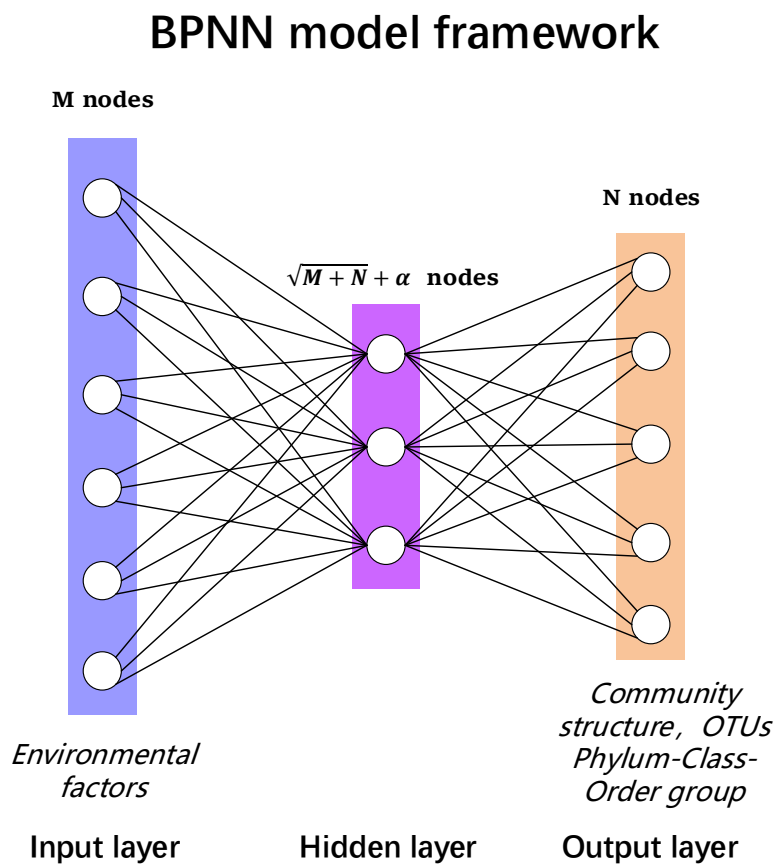

**Figure S1.** The framework of BP models: input data (blue), output data (red), and a hidden layer compute output data from input data (purple)

**Supplementary Note 1: Detailed description of the ANN and BPNN method.**

Artificial neural network (ANN) [36] is a machine learning method that automatically and quantitatively learns relationships without needing specific assumptions or guided system optimization. Artificial neural networks are an excellent tool for modeling complex connections between microbial communities and environmental variables, especially because of their ability to handle non-linear relationships and interactions between predictor variables. In a variety of ecosystems, artificial neural networks have enabled researchers to successfully study the correlation between environmental factors and microbial community structure. However, this method is not widely applied in activated sludge systems. The Back Propagation Neural Network (BPNN) [37-38] is a machine learning approach that allows for the automatic and quantitative learning of relationships without requiring specific assumptions or guided system optimization. The BPNN employs forward propagation of data and backward propagation of errors, utilizing the gradient descent algorithm to update weight thresholds, thus completing the training process. Compared to conventional machine learning approaches [39-41], the BPNN exhibits superior capabilities in non-linear modeling, representation of high-dimensional feature spaces, ease in handling time series tasks, and the flexibility of end-to-end learning.

As depicted in **Figure S1**, once initial parameters are chosen, the gradient descent is an excellent tool to model the intricate connections between microbial communities and environmental variables, particularly due to its capacity to handle non-linear relationships and interactions among predictors. In various ecosystems, BPNNs have enabled researchers to investigate the correlation between environmental factors and microbial community structure [42-44]. However, applications of this method in activated sludge systems are still insufficient. Furthermore, while the convergence of the gradient descent algorithm in Backpropagation Neural Networks (BPNN) has been theoretically established, it is crucial to acknowledge that the convergence point may not always correspond to the global optimum. Additionally, the selection of initial parameters significantly influences the convergence behavior. The random selection of initial parameters may not consistently yield an optimal starting point, and the reliability and stability of the model are profoundly dependent on the initial parameter configuration.

**Supplementary Figure S2: The operating strategy of the differential evolution algorithm.**

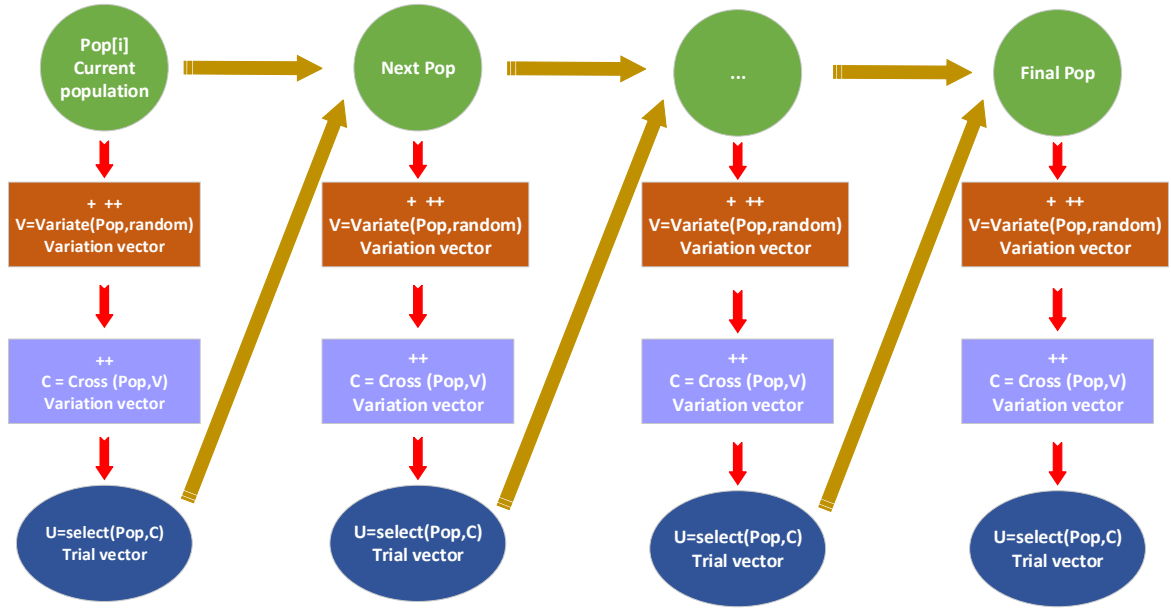

**Figure S2.** The operating strategy of the differential evolution algorithm

**Supplementary Note 2: Detailed description of the statistical analysis method.**

In contrast to gradient descent, the differential evolution (DE) algorithm, proposed by Storn et al. [45], is a heuristic algorithm known for its excellent global search capability. Differential evolution algorithms encompass fundamental operations, including mutation [46], crossover [47], and selection [48]. As depicted in **Figure S2**, differential evolution executes these operations on the initial population to identify the most promising individuals. Swarm intelligence emerges from the interplay of cooperation and competition among individuals, directing the course of optimization. Therefore, the DE algorithm can be introduced into BPNN to address the initial parameter values optimization problem. In the DE-BP neural network, parameters are treated as decision variables, and the model's accuracy serves as the objective function. The algorithmic process of optimizing the BP neural network with the differential evolution algorithm (DE) is depicted in **Figure S3**.

**Supplementary Figure S3: Schematic diagram of the DE-BP neural network.**

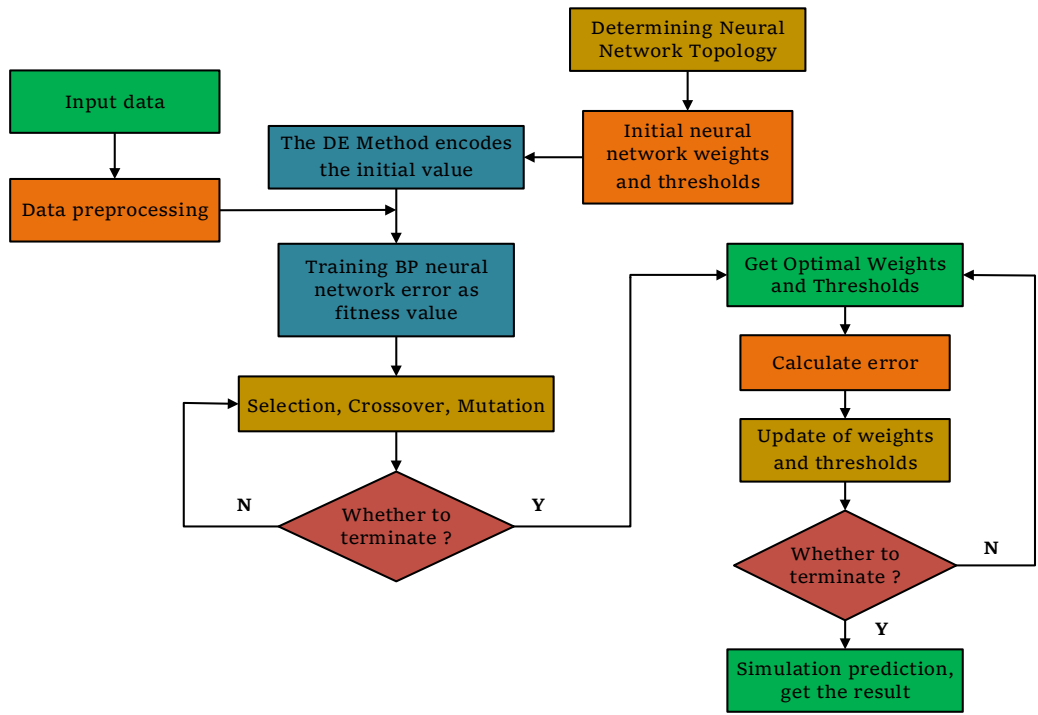

**Figure S3.** The flowchart of BP neural network based on differential evolution

**Supplementary Note 3: The DE-BP neural network for WWTPs prediction.**

Here, we utilized the DE-BP neural network [49-50] and the data of environmental and geographical, influent, effluent, and reactor operation to predict the microbial community structure of AS systems from global WWTPs. The DE-BP neural network utilized in this topic is structured as 37-8-21, 37-9-51, and 37-15-171 at Phylum, Class, and Order levels, respectively. The network takes 37 features from WWTPs as inputs and predicts the microbial composition as outputs, categorized into 21, 51 and 171 classes at the Phylum, Class, and Order levels, respectively. We compared the predicted microbial community structure and convergence speed between the DE-BP neural network model and the BPNN model across three levels: Phylum, Class, and Order. Then, we analyzed the impact of various WWTP characteristic factors on the predictions, and the predictive accuracy was measured relative to the 1:1 observed-predicted normal distribution. These analyses have deepened our understanding of the microbial communities in AS systems, providing recommendations for accurately predicting main functional groups and offering a theoretical basis for improved design and operational parameter adjustments as well as controlling community structures. In addition, compared to conventional BP neural network-based predictions, the DE-BP approach exhibits superior convergence accuracy and results in smaller prediction errors.

**Supplementary Figure S4: The abstract representation of clustering method.**

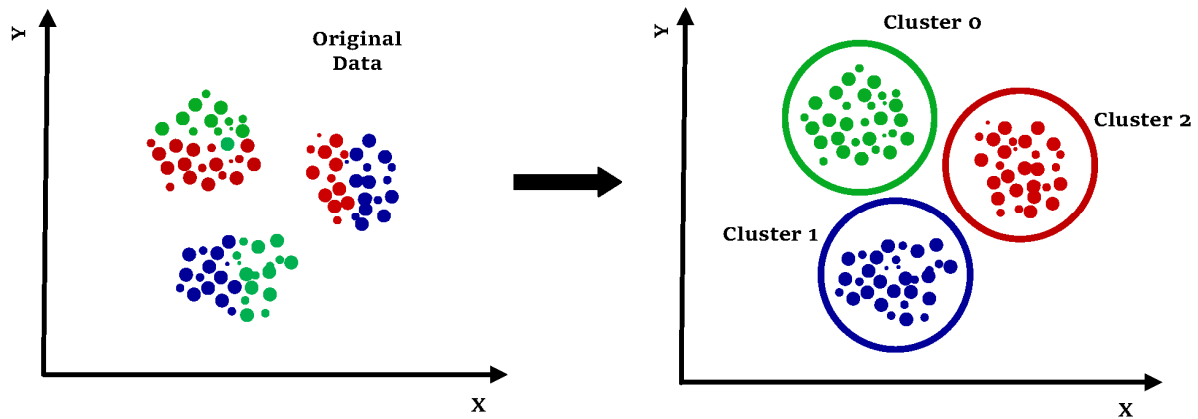

**Figure S4.** Abstract representation of clustering method

**Supplementary Note 4: Detailed description of the WWTPs clustering.**

For the clustering analysis of global WWTPs under various feature attributes, we utilized various feature attributes of WWTPs, including influent features, effluent features, reactor operational features, and environmental and geographical features. Ultimately, this approach enables the clustering analysis of similar WWTPs based on different feature attributes. It allows us to unveil the microbial composition at the Phylum, Class, and Order level and then generate heatmaps depicting the proportion of different microbial communities among these WWTPs. Clustering is a data technique that divides a collection of abstract or physical objects into distinct groups of similar items. As shown in **Figure S4**, the primary purpose of clustering is to maximize the similarity among data points within the same group while minimizing the similarity between different groups. This process enables the identification of meaningful patterns and structures within the dataset, aiding in data analysis and decision-making.

**Supplementary Figure S5: Schematic diagram of the DPNG-EPMC model.**

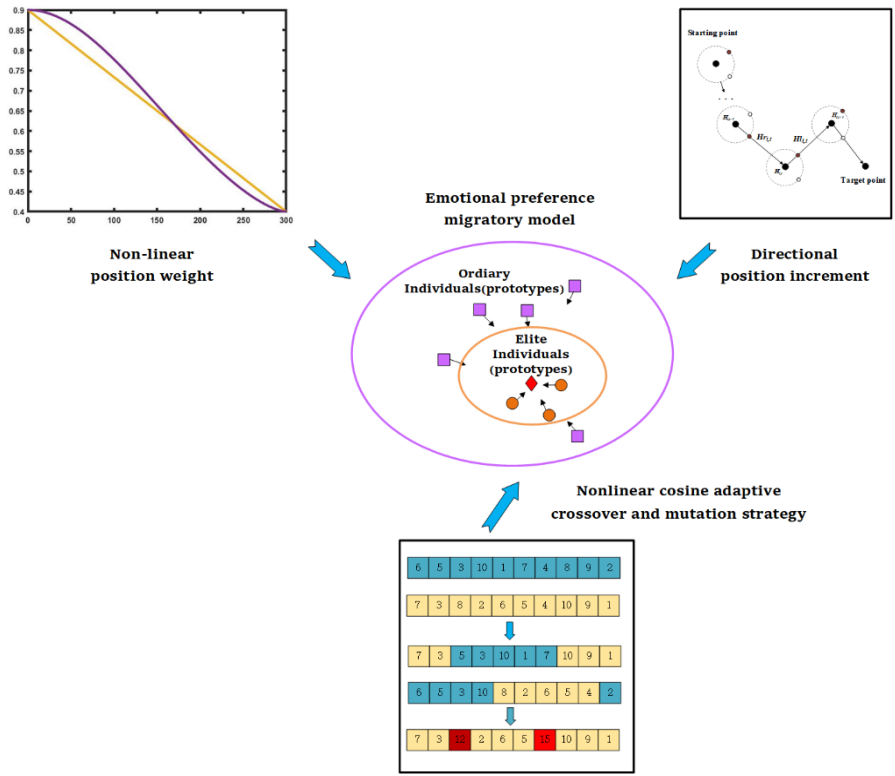

**Figure S5.** The relationship between the sub-models of DPNG-EPMC

**Supplementary Note 5: Detailed description of the DPNG-EPMC model clustering.**

The Emotional Preference and Migratory Behavior (EPMC) model [51] has been proposed as an effective Swarm intelligence approach for dealing with clustering tasks, which consists of four essential models: the migration model, the emotional preference model, the social group model, and the inertial learning model. While the EPMC model has demonstrated efficiency in addressing clustering tasks, it is not without drawbacks, namely premature convergence and a lack of population diversity. In response to these challenges, we have integrated a range of evolutionary strategies into the EPMC clustering model. These strategies encompass the non-linear position weight, directional position increment, and a novel non-linear cosine adaptive crossover and mutation method. This concerted effort has led to the introduction of a new emotional preference migration clustering model termed DPNG-EPMC. **Figure S5** illustrates the main architecture and relationship between each sub-model of the DPNG-EPMC. Through its design, DPNG-EPMC effectively safeguards against premature convergence to local optima, enhances population diversity, and enriches the spectrum of available clustering methodologies.

**Supplementary Figure S6: Schematic diagram of the SiTime-GAN model.**

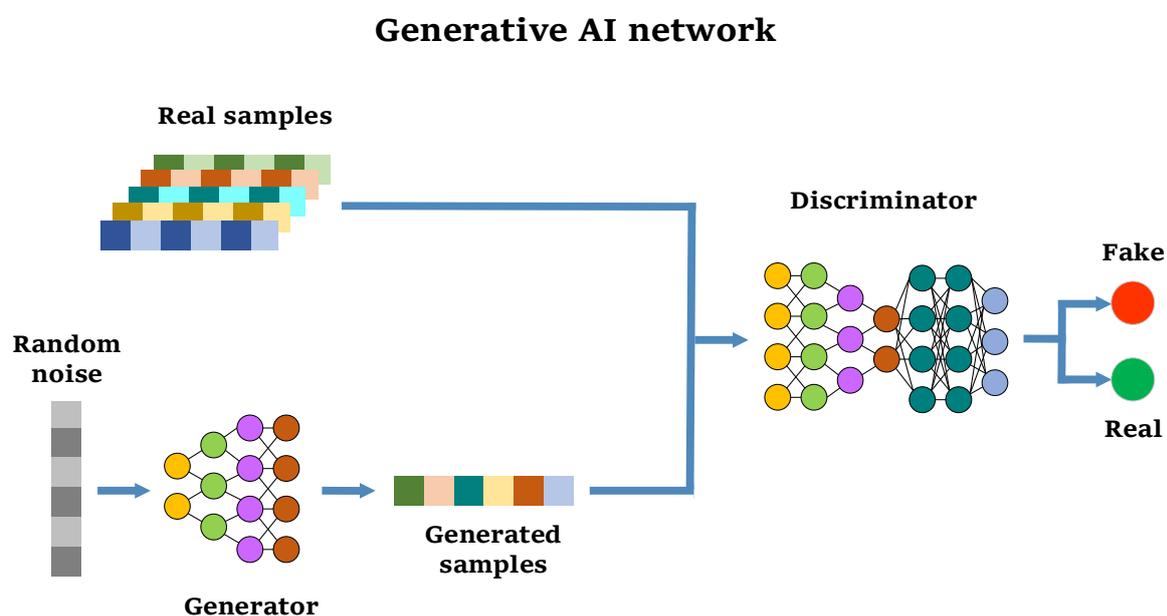

**Figure S6.** The Generative AI for wastewater treatment plants (WWTPs) microbial composition prediction

**Supplementary Note 6: Detailed description of the SiTime-GAN model.**

For the data generation of environmental and geographical features, the influent feature, the effluent feature, and the reactor operation feature to facilitate the prediction of the microbial community structure, we employ a model similar to Time-GAN [52] to generate feature data for different attributes of WWTPs. Without considering the time sequence dimension, we use the Similar Generative Adversarial Networks (SiTime-GAN) model to generate data that includes various feature attributes and microbial community compositions of WWTPs at the Phylum, Class, and Order levels. SiTime-GAN leverages the competition between two neural networks to learn the probability distribution of a dataset. One of these networks, named generator $G$, creates novel data instances [53], while the other, the discriminator $D$, evaluates their authenticity [54]. The Generative AI for wastewater treatment plants (WWTPs) microbial composition prediction is shown in **Figure S6**.

As shown in **Figure S6**, firstly, the generator takes in random numbers and produces generated data; then, the generated data, along with data obtained from an actual real dataset, is fed into the discriminator. Finally, the discriminator receives both real and generated data and outputs probabilities (numbers between 0 and 1), where 1 represents a prediction of the data's authenticity, and 0 represents falseness. The discriminator $D$ updates its parameters based on the loss, $G$ distinguishing real data as true and generated data as false. Meanwhile, generator G updates its parameters through loss to have its generated data considered true. Eventually, $G$ becomes a flawless generator, producing data that confounds the discriminator $D$ into being unable to differentiate between real and generated data. This state is known as the Nash equilibrium, where the capabilities of the generator $G$ and the discriminator $D$ are honed to perfection through a game, rendering them equally proficient.

Based on the aforementioned theory, we can apply it to the generation of raw data for wastewater treatment plant (WWTPs) microbial composition prediction. Here, we utilize the SiTime-GAN model to generate feature data for various feature attributes and microbial community compositions of WWTPs at the Phylum, Class, and Order levels. We aim to generate high-quality feature attribute data with the same dimensions resembling real data. Finally, by generating substantial data on different attributes and microbial community composition of WWTPs, we can subsequently employ the newly generated data to train the DE-BP model further, aiming for enhanced solutions to pertinent scientific needs and research objectives.

**Supplementary Figure S7: The comparison of the predictive performance of DE-BP and BPNN model for more Phylum, Class and Order in microbiome community.**

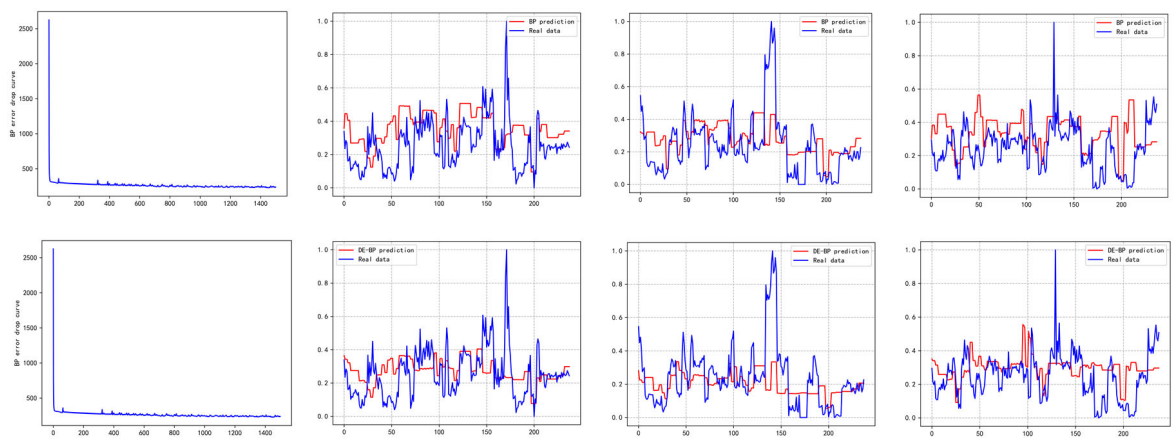

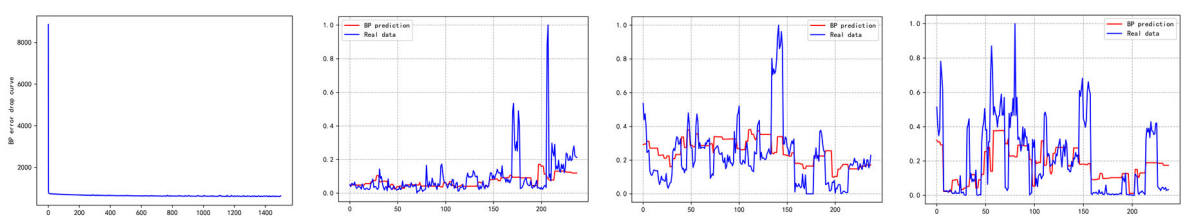

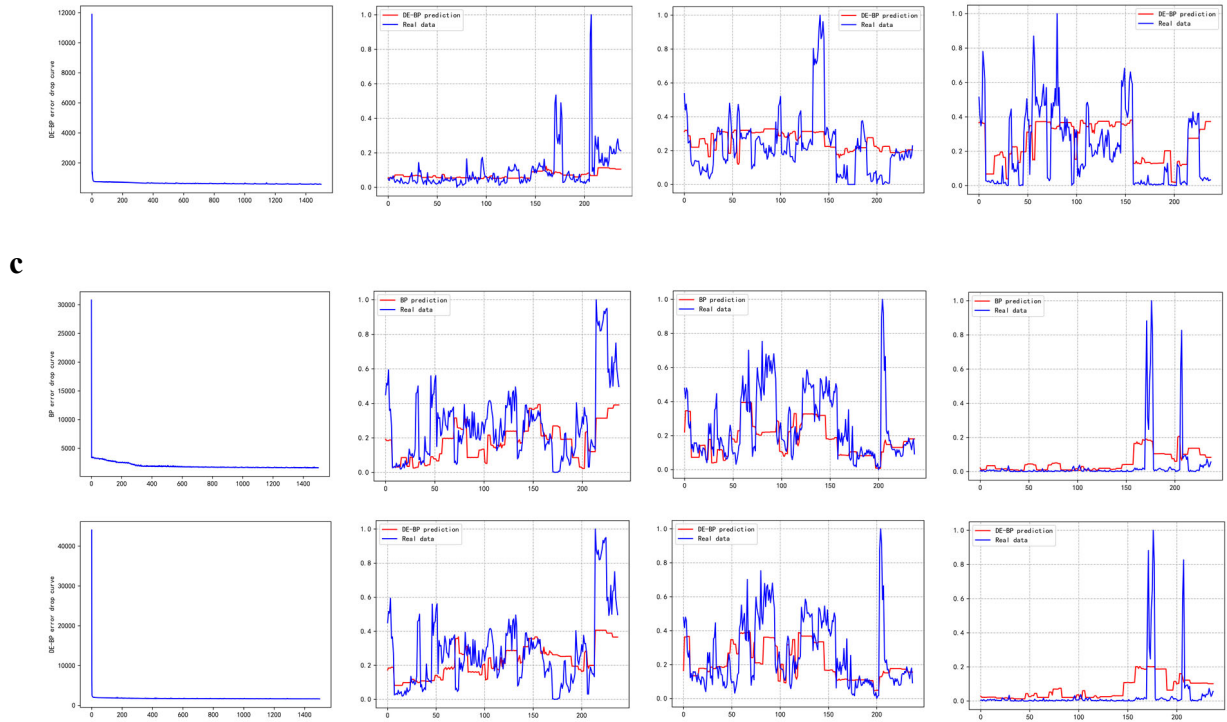

**Figure S7.** The comparison of the predictive performance of DE-BP and BPNN model for microbiome community structure. a, The error drop curves for one epoch and the comparison between the predicted values and observed values across the test dataset for BPNN and DE-BP models on the Phylum level. b, The error drop curves for one epoch and the comparison between the predicted values and observed values across the test dataset for BPNN and DE-BP models on the Class level. c, The error drop curves for one epoch and the comparison between the predicted values and observed values across the test dataset for BPNN and DE-BP models on the Order level.